\documentclass[lettersize,journal]{IEEEtran}

\usepackage{amsmath,amsfonts,bm}

\def\Figref#1{Figure~\ref{#1}}

\def\Tabref#1{Table~\ref{#1}}

\def\Secref#1{Section~\ref{#1}}

\def\Eqref#1{Equation~\eqref{#1}}

\def\1{\bm{1}}

\DeclareMathAlphabet{\mathsfit}{\encodingdefault}{\sfdefault}{m}{sl}
\SetMathAlphabet{\mathsfit}{bold}{\encodingdefault}{\sfdefault}{bx}{n}

\newcommand{\firstres}[1]{{\textbf{\textcolor{red}{#1}}}}
\newcommand{\secondres}[1]{{\underline{\textcolor{blue}{#1}}}}
\newcommand{\thirdres}[1]{{\textbf{\textcolor[HTML]{008000}{#1}}}}

\usepackage{textcomp}
\usepackage{stfloats}
\usepackage{verbatim}

\usepackage{tabularx,booktabs}
\usepackage{color}
\usepackage[x11names]{xcolor}
\usepackage{tabularray}
\UseTblrLibrary{diagbox}
\usepackage{threeparttable}
\usepackage{multirow}
\usepackage{cite}

\usepackage{amsmath}
\usepackage{amsfonts}
\usepackage{amssymb}
\usepackage{extarrows}
\usepackage{amsthm}
\usepackage[linesnumbered,ruled,lined]{algorithm2e}
\usepackage{array}
\usepackage{multirow}
\usepackage{makecell,rotating}
\usepackage{float}

\usepackage{subfigure}
\usepackage{graphicx}
\graphicspath{ {./figures/} }

\usepackage{url}
\usepackage[colorlinks,linkcolor=blue,anchorcolor=blue,citecolor=blue]{hyperref}
\usepackage{orcidlink}

\newcommand{\posres}[1]{\textcolor{red}{#1}}
\newcommand{\negres}[1]{\textcolor{blue}{#1}}

\begin{document}

\title{Time Evidence Fusion Network: Multi-source View in Long-Term Time Series Forecasting}

\author{Tianxiang Zhan$^{\orcidlink{0000-0002-5188-0381}}$,
    Yuanpeng He$^{\orcidlink{0000-0001-9071-2260}}$,
    Yong Deng$^{\orcidlink{0000-0001-9286-2123}}$,
    Zhen Li$^{\orcidlink{0000-0002-5156-3730}}$,
    Wenjie Du$^{\orcidlink{0000-0003-3046-7835}}$,
    Qingsong Wen$^{\orcidlink{0000-0003-4516-2524}}$
    \thanks{Code is available at this repository: \url{https://github.com/ztxtech/Time-Evidence-Fusion-Network}.}
    \thanks{This work was supported in part by the National Natural Science Foundation of China under Grant 62373078.}
    \thanks{Tianxiang Zhan is with Institute of Fundamental and Frontier Science, University of Electronic Science and Technology of China, Chengdu 610054, China and also with PyPOTS Research (e-mail: ztxtech@std.uestc.edu.cn).}
    \thanks{Yuanpeng He is with the Key Laboratory of High Confidence Software Technologies (Peking University), Ministry of Education, Beijing, 100871, China; School of Computer Science, Peking University, Beijing, 100871, China (e-mail: heyuanpeng@stu.pku.edu.cn).}
    \thanks{Yong Deng is with Institute of Fundamental and Frontier Science, University of Electronic Science and Technology of China, Chengdu 610054, China (e-mail: dengentropy@uestc.edu.cn).}
    \thanks{Zhen Li is with the China Mobile Information Technology Center, Beijing 100029, China (e-mail: zhen.li@pku.edu.cn).}
    \thanks{Wenjie Du is with PyPOTS Research (e-mail: wdu@pypots.com).}
    \thanks{Qingsong Wen is with Squirrel Ai Learning, Bellevue, USA (e-mail:  qingsongedu@gmail.com).}
    \thanks{Corresponding author: \textit{Yong Deng}}
}

\markboth{Journal of \LaTeX\ Class Files,~Vol.~14, No.~8, August~2021}%
{Shell \MakeLowercase{\textit{et al.}}: A Sample Article Using IEEEtran.cls for IEEE Journals}

\maketitle

\begin{abstract}
    In practical scenarios, time series forecasting necessitates not only accuracy but also efficiency. Consequently, the exploration of model architectures remains a perennially trending topic in research. To address these challenges, we propose a novel backbone architecture named Time Evidence Fusion Network (TEFN) from the perspective of information fusion. Specifically, we introduce the Basic Probability Assignment (BPA) Module based on evidence theory to capture the uncertainty of multivariate time series data from both channel and time dimensions. Additionally, we develop a novel multi-source information fusion method to effectively integrate the two distinct dimensions from BPA output, leading to improved forecasting accuracy. Lastly, we conduct extensive experiments to demonstrate that TEFN achieves performance comparable to state-of-the-art methods while maintaining significantly lower complexity and reduced training time. Also, our experiments show that TEFN exhibits high robustness, with minimal error fluctuations during hyperparameter selection. Furthermore, due to the fact that BPA is derived from fuzzy theory, TEFN offers a high degree of interpretability. Therefore, the proposed TEFN balances accuracy, efficiency, stability, and interpretability, making it a desirable solution for time series forecasting.
\end{abstract}


\section{Introduction}
\IEEEPARstart{T}{he} evolution of various phenomena is inherently tied to the progression of time, leading to an increasing prevalence of time series data in a multitude of application domains \cite{wen2022robust} such as medical diagnosis \cite{al2020adaptive, fan2021detection,li2023causal}, education \cite{mao2024time}, weather \cite{wang2020deep, zhang2023skilful}, power system \cite{qin2023personalized}, and so on \cite{du2023pypots, wu2023copp, li2024opf}. The importance of time series forecasting cannot be overstated. There are many methods for time series forecasting. The initial methods came from statistics, such as Autoregressive integrated moving average (ARIMA) model \cite{hyndman2018forecasting}. Deep learning is a powerful method with wide applications in various disciplines  \cite{hou2024dag, zhang2024multivariate}. In recent years, time series prediction models based on neural networks have become a research hotspot. These neural networks have evolved from the simplest deep neural networks (DNN) \cite{bengio2017deep}. Until now, there are already many backbones that can be used for time series, such as Recurrent Neural Network (RNN) \cite{lu2024trnn}, Long Short-Term Memory (LSTM) \cite{yadav2024noa}, Transformer \cite{wen2023transformers, Informer, Crossformer, wang2024card}, and so on \cite{jin2024survey,shao2024exploring}. Large Language Models (LLM) for time series also perform well in cross dataset forecasting tasks \cite{das2023decoder, jintime, liang2024foundation}.

\begin{figure*}[t]
    \centering
    \includegraphics[width=0.8\linewidth]{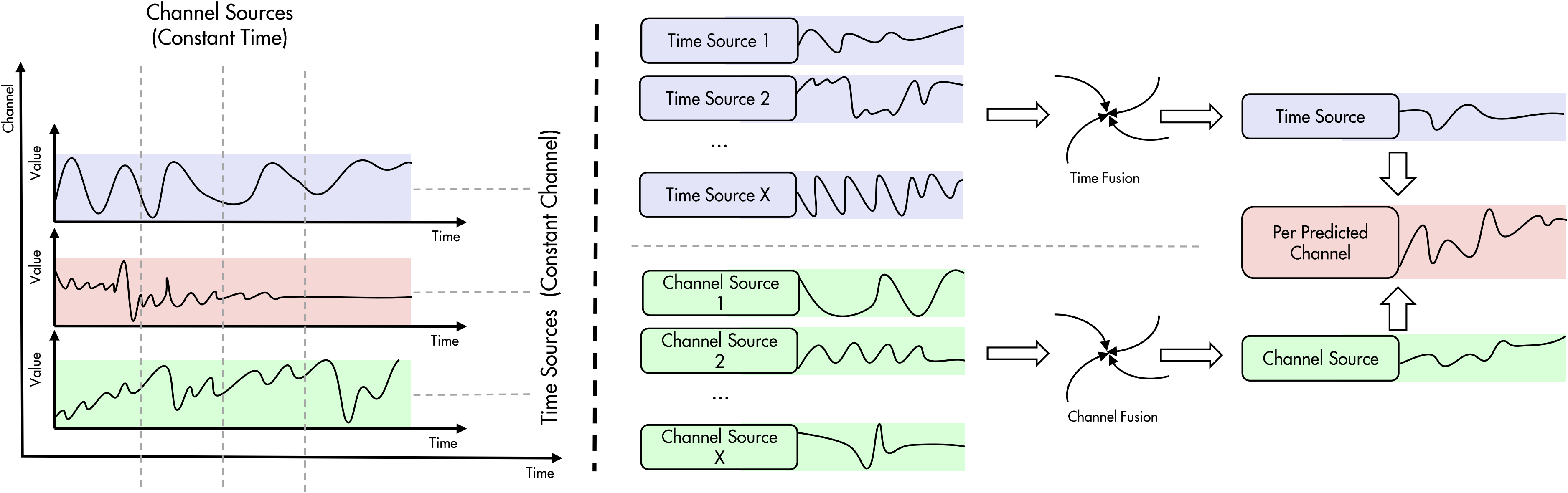}
    \caption{The channel dimension and time dimension of multivariate time series can be considered as different sources of information. Firstly, merge information sources of the same type, and finally integrate information sources from both channel and time dimensions to complete the final prediction.}
    \label{fig:ms}
\end{figure*}

There is uncertainty in time series data, and modeling uncertainty on time series is effective. Evidence theory is an uncertain reasoning framework based on mass function \cite{DS1, DS2, yang2013evidential, deng2024random}. Previous studies have shown that evidence theory can effectively utilize the uncertainty of time series data and achieve positive application results \cite{zeng2023new, zhang2024belief}. On the other hand, evidence theory is also a method of information fusion that can effectively aggregate different information sources and make effective decisions \cite{smets1990combination, smets1994transferable}. Meanwhile, evidence theory can also be applied in neural networks to accomplish tasks such as classification \cite{denoeux2000neural}. Furthermore, the essence of evidence theory lies in using mass functions to model variables. A mass function, as a generalized form of probability, can directly represent the uncertainty among different arguments that is challenging to achieve with traditional probability models. Therefore, in order to reduce the error of time series forecasting, we propose the Time Evidence Fusion Network (TEFN) from the perspectives of evidence theory and information fusion. TEFN is a lightweight neural network that considers the time and channel dimensions of multivariate time series as different sources of information in \Figref{fig:ms}.

\begin{figure}[h]
    \centering
    \includegraphics[width=0.8\linewidth]{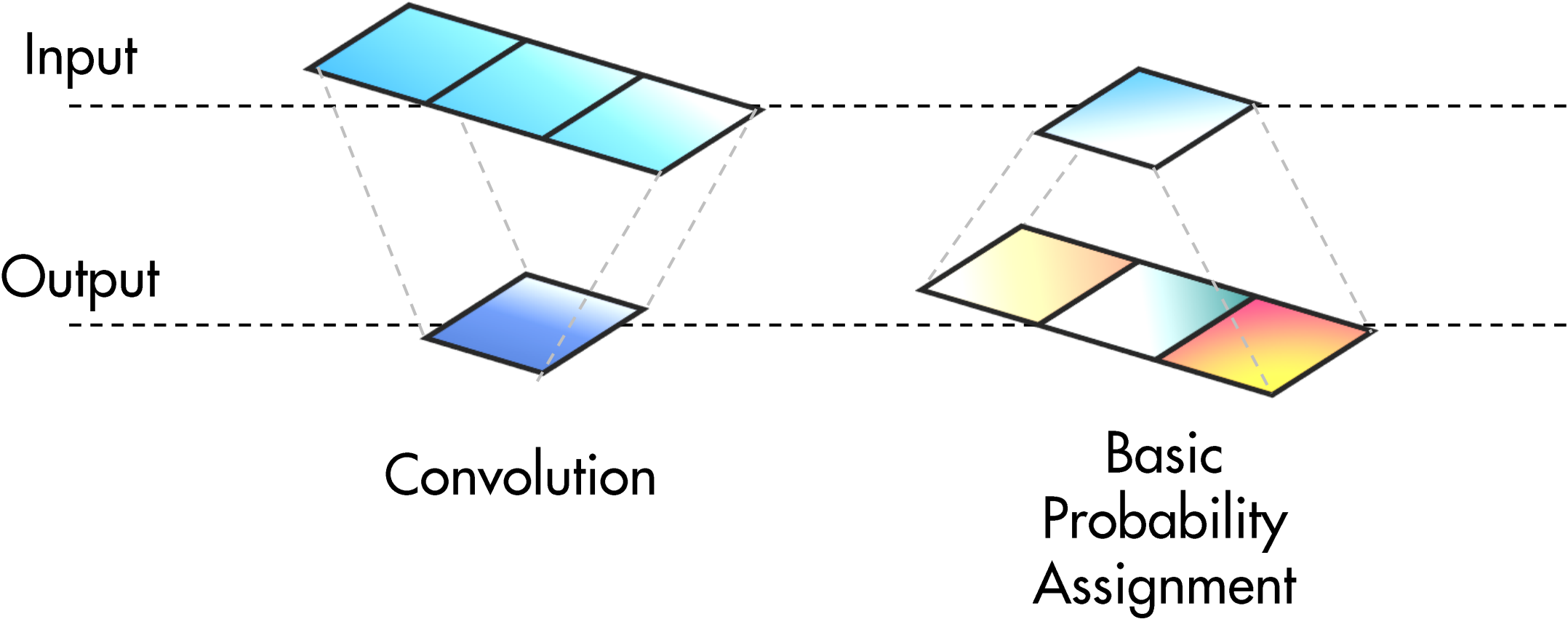}
    \caption{Comparison between convolution and basic probability assignment: convolution condenses information while basic probability assignment diverges to consider different possibilities.}
    \label{fig:inver_cov}
\end{figure}

In TEFN, we propose a novel Basic Probability Assignment (BPA) Module. Based on the evidence theory of decision-making process, BPA maps different sources of information to distributions related to the target outcome. The target distribution is achieved by the fuzzy membership of different fuzzy sets \cite{deng2015generalized}. In image processing, convolution condenses the required features of the target \cite{bengio1993globally}. Unlike convolution, BPA is an expansion process in \Figref{fig:inver_cov}. On the contrary, BPA considers all possibilities in the event space composed of subsets of the sample space, expanding on different events, which is more effective for simple structures like time series and can fully explore the different hidden information inside. For different distributions generated by multiple information sources, we consider sources as samples and use the expected forecasting values as the final result.

In this article, we focus on the practical application details of TEFN, including its accuracy, memory usage, training speed, parameter sensitivity, and interpretability. At the same time, in order to ensure fairness, the experimental data and configuration files for comparison are all from open-source projects. TEFN achieves state-of-the-art (SOTA) on multiple real datasets, while training time and model parameters are much smaller than models based on Transformer \cite{Crossformer, PatchTST}. For random hyperparameter selection, TEFN exhibits weak fluctuations and is a stable model. TEFN holds significant value and potential impact in various practical application scenarios. In the energy sector, such as power systems, accurate electricity consumption forecasting is crucial for the reliable operation, management, and planning of the power grid \cite{koot2021usage,zhu2023energy}. With large scale power consumption data, traditional forecasting methods may face high computational complexity and long training times. TEFN, however, can efficiently process millions of time series, enabling power companies to optimize power generation and distribution, reducing energy waste and costs. In the context of AIOps, resource forecasting and optimization require accurate predictions to ensure the stable operation of cloud systems \cite{zhou2023ahpa}. As cloud platforms manage millions of time series data related to resource usage, low complexity and high efficiency forecasting methods are essential \cite{guo2024pass}. TEFN can provide precise resource usage forecasts, helping cloud operators to allocate resources effectively, reduce operational costs, and improve service reliability.

The contributions of this article are as follows:
\begin{itemize}
    \item A novel evidential backbone BPA designed for extracting potential distributions from simple data structures is introduced.
    \item A BPA-based neural network, TEFN, is introduced for time series forecasting, outpacing prior research in terms of accuracy, efficiency, stability, and interpretability.
    \item The manuscript elucidates the incorporation of evidence theory concepts, such as the mass function, into the architecture of neural networks, offering a novel approach to uncertainty reasoning in machine learning.
\end{itemize}

This paper is organized as follows. We begin by reviewing relevant literature in the \Secref{sec:rw}. Next, in the \Secref{sec:tefn}, we delve into the architecture and underlying mathematical principles of TEFN. In the \Secref{sec:exp}, we present a series of experiments that demonstrate TEFN's superior performance. Finally, we conclude the paper by summarizing our key findings in the \Secref{sec:con}.

\section{Related Work}\label{sec:rw}
\subsection{Dempster-Shafer Theory} \label{sec:ds}
Dempster-Shafer theory, also known as Evidence Theory, operates on a different framework compared to traditional probability theory \cite{DS1,DS2}. It relies on the concept of mass functions, which provide a more flexible representation of uncertainty, accommodating weaker constraints than traditional probabilities. All independent and mutually exclusive sample sets that exist in the system are called frames of discernment (FOD), represented as $\Omega$.

When there are multiple sources of information, information fusion is usually required to make decisions. Fusing two mass functions, denoted as $m_1$ and $m_2$, the fused mass function $m$ is determined using the Dempster-Shafer Rule (DSR), as described in \Eqref{eq:ds}. The symbol variables $A$, $B$, and $C$ are focal elements, which are typically combinations of all independent samples in the FOD, which are often composed of target labels, such as the classification targets.

\begin{equation}
    m(A) = \sum_{B \cap C =A} m_1(B)*m_2(C)
    \label{eq:ds}
\end{equation}

Here is an example of an intuitive explanation of the mass function. Suppose there are two unidentifiable features $A$ and $B$ in the time series. There are different ways $1$ and $2$ to identify whether it is $A$ or $B$ (For example, whether the increase in time series values is due to a trend or a cycle.). Therefore, we obtain the results in in Equation \eqref{eq:ds_example}, and both ways are able to distinguish the feature of $A$, but there is also a certain degree that couldn't distinguish between $A$ and $B$. It should be noted that if modeling is done using probabilities, the inability to distinguish between $A,B$ cannot be directly expressed. The usual approach is to declare a new argument $C$ as indistinguishable between $A$ and $B$. But at this time, $C$ will be independent of $A$ and $B$, and the connection between propositions will be lost. DSR can unify the opinions of the two methods $m_1, m_2$ and obtain a comprehensive opinion $m$ in Equation \eqref{eq:ds_example}.

\begin{equation}
    \begin{split}
         & \text{Mass 1: }m_1(A) = 0.7 \quad m_2(A, B)=0.2 \\
         & \text{Mass 2: }m_2(A) = 0.4 \quad m_2(A,B)=0.7  \\
         & \text{Fused Mass: }m(A)=0.86 \quad m(A,B)=0.14  \\
    \end{split}
    \label{eq:ds_example}
\end{equation}

\subsection{Basic Probability Assignment}\label{sec:bpa}
BPA is a core concept in evidence theory, used to map real data $x$ into mass distributions. Mass distributions describe the degree of support for various hypotheses in the event space, providing a basis for reasoning and decision-making.

The generation of BPA $m$ is often based on fuzzy logic, where fuzzy membership functions $\mu$ are used to map data points to different fuzzy sets, thereby constructing the target distribution in \Eqref{eq:fuzzy} where $w$ and $b$ is weight and bias of the membership function. This kind of membership is called the triangular membership function, assuming that the membership is linearly related to the random variable.

\begin{equation}\label{eq:fuzzy}
    \begin{split}
        m = \mu(x) = wx+b
    \end{split}
\end{equation}

\begin{figure}[h]
    \centering
    \includegraphics[width=0.8\linewidth]{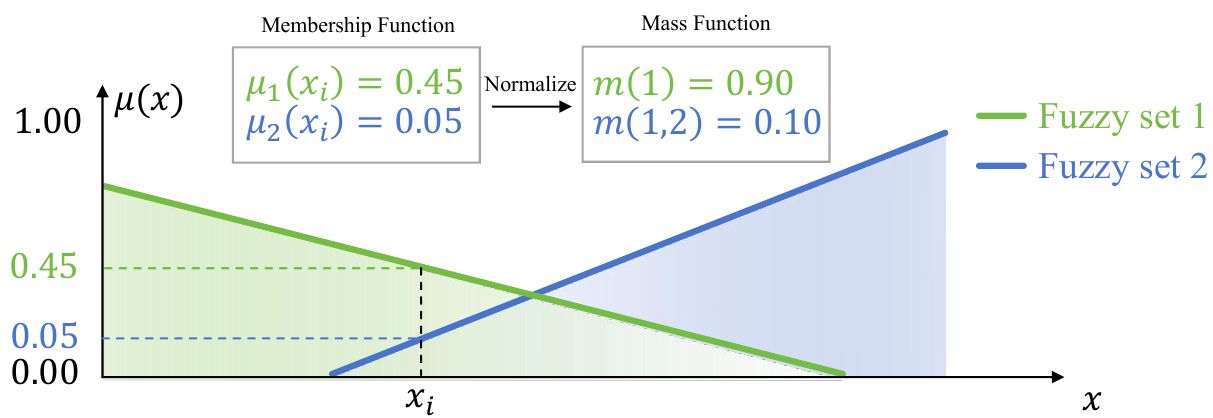}
    \caption{The conversion between fuzzy membership and mass function.}
    \label{fig:fuzzy_intro}
\end{figure}

An intuitive example is shown in Figure \ref{fig:fuzzy_intro}. There are two fuzzy sets, which are assumed to represent the categories of random variables. We randomly take a sample $x_i$, and at this time, the membership function reflects the degree to which $x_i$ belongs to this fuzzy set, which can be understood as the degree of belonging to this category. This is also different from probability. The membership degree is bounded and can only reflect the extent to which it belongs to the current fuzzy set. Therefore, when multiple fuzzy sets appear, they need to be normalized in order to be converted into a probability distribution or a mass function. And the mass function can increase the description of uncertainty. In Figure \ref{fig:fuzzy_intro}, since the degree of belonging to the first category is greater, we believe that the degree of belonging to the second category actually reflects that the two fuzzy sets are not clearly distinguishable.

BPA can also be generated by assigning support degrees through Gaussian distributions $N$ (taking probability density function as a membership function $f$) where $\mu$ and $\sigma$ are expectation and variance in \Eqref{eq:x_g1}.

\begin{equation}\label{eq:x_g1}
    \begin{split}
         & x \sim N(\mu,\sigma)                                                     \\
         & m = f(x) = \frac{1}{\sqrt{2\pi\sigma^2}}e^{-\frac{(x-\mu)^2}{2\sigma^2}} \\
    \end{split}
\end{equation}

The advantage of BPA lies in its ability to effectively handle uncertainty and improve the accuracy and reliability of predictions by fusing multi-source information. On the one hand, the existence of fuzzy sets allows elements to partially belong to a set, which allows for a more continuous attribution, rather than the set only belonging to this set or not belonging to this set.

\subsection{Evidence Decision Making}
The process of Evidence Decision Making can be delineated into several sequential steps: basic probability assignment (BPA), evidence fusion, and decision making (DM) \cite{deng2015generalized, zhan2024random}. BPA serves as a pivotal initial phase, constituting a mapping from raw data to mass distribution, akin to the path from raw data to probability distribution. Subsequently, evidence fusion integrates distinct mass distributions from various data channels, culminating in a rationalized mass distribution. The most classic fusion method is the DSR \cite{DS1,DS2} mentioned in \Secref{sec:ds}. Finally, decision making transpires as the progression from the attained distribution towards the designated target. The most classic decision-making method is to transform the mass distribution into the probability distribution of the target, called pignistic probability transformation (PPT) \cite{dubois1993possibility} in \Eqref{eq:ppt}, where $x$ is a target label and $e$ are combinations of all possible target labels.

\begin{equation}\label{eq:ppt}
    p(x) = \sum_{x \in e} \frac{m(e)}{|e|}
\end{equation}

\subsection{Linear Time Series Model}
There are many linear time series forecasting models (LTSFM) \cite{hyndman2018forecasting, DLinear, LightTS, shao2022spatial}. LTSFM offer several advantages for modeling and forecasting time series data:
\begin{itemize}
    \item The simplicity and transparency make them easy to implement, interpret, and analyze.
    \item This leads to faster training and inference times, making them suitable for real-time applications and large datasets.
    \item No additional encoding layers are required.
\end{itemize}

A classic LTSFM model is DLinear \cite{DLinear}. DLinear serves as a simple yet effective baseline for long-term time series forecasting. It is a model that employs a seasonal-trend decomposition technique to separate the trend and seasonal components of the time series data. Each component is then independently modeled using a single-layer linear regression. The final prediction is obtained by summing the outputs of the two linear regressions.

\section{Time Evidence Fusion Network}\label{sec:tefn}

\begin{figure}[t]
    \centering
    \includegraphics[width=0.6\linewidth]{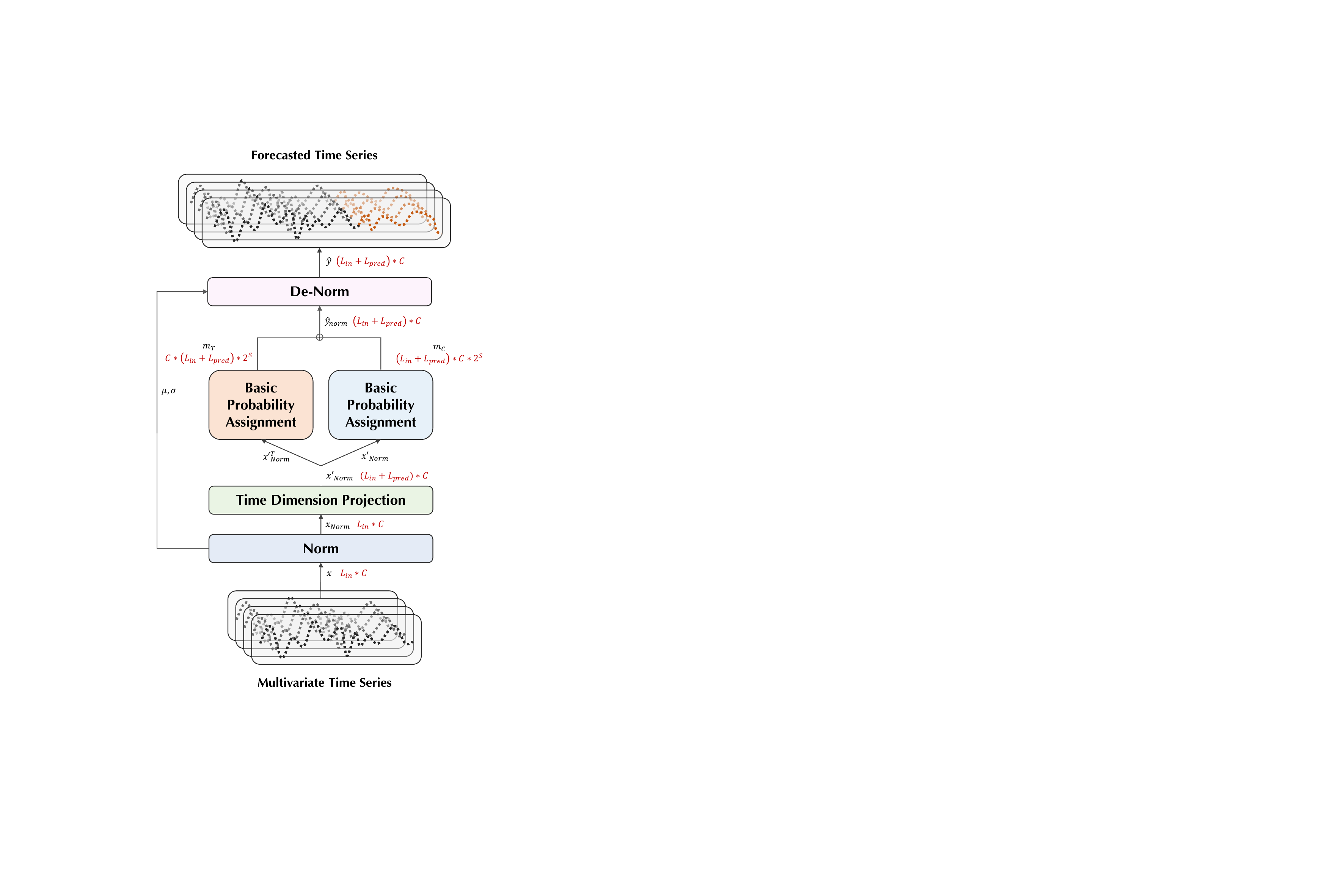}
    \caption{The overall structure of TEFN: Establish corresponding mass functions for the time dimension and channel dimension as the target dimensions of BPA. }
    \label{fig:tefn}
\end{figure}

The structure of TEFN is inherently straightforward, achieving remarkable performance with a design featuring only two BPA modules, akin to employing just two convolutional kernels. We posit that the novel architecture of TEFN embodies the essence of a pure backbone model, stripped of domain-specific intricacies. This minimalist approach allows TEFN to focus solely on capturing the fundamental temporal patterns within the data, making it a versatile and adaptable foundation for various time series analysis tasks. The overall structure of TEFN is depicted in \Figref{fig:tefn}. Next, we will elaborate on each module one by one.

\begin{figure*}[t]
    \centering
    \includegraphics[width=0.6\linewidth]{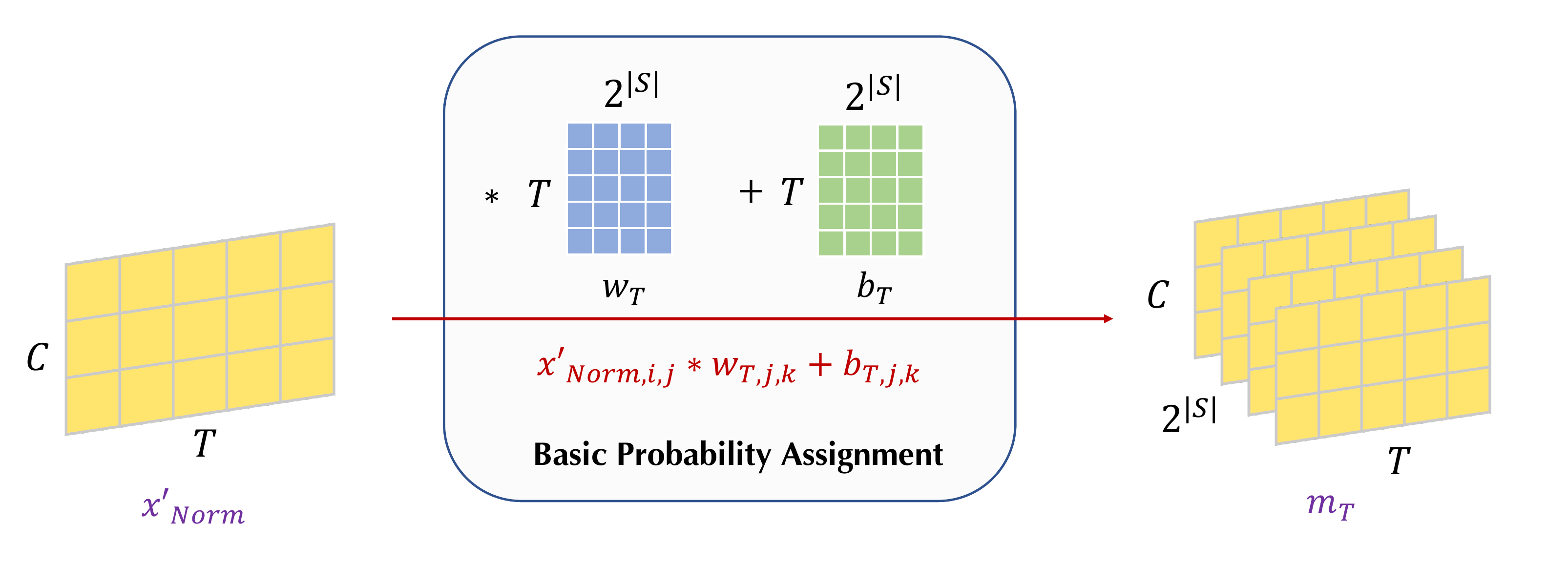}
    \caption{The operation of a BPA module: weight matrix and bias matrix are both learnable parameters. BPA elevates the dimensionality of data by adding a dimension to distinguish different events.}
    \label{fig:bpa}
\end{figure*}

\subsection{Time Normalization and De-normalization}\label{sec:model}

To expedite convergence towards local optima and enhance training efficiency, TEFN incorporates a normalization technique inspired by the Stationary model \cite{Stationary}. This process computes the mean $\mu$ as shown in \Eqref{eq:norm_mean} and variance $\sigma^2$ as shown in \Eqref{eq:norm_var} for each segment of the time series $x$. The subsequent normalization and de-normalization operations, presented in \Eqref{eq:norm} and \Eqref{eq:denorm}, where $\hat y$ is forecasting result, respectively, act as symmetrical transformations that standardize the time series data. The neural network operation $Net(\cdot)$ is then applied to the normalized time series $x_{Norm}$, ensuring that the model learns from data with consistent scale and distribution. This normalization strategy effectively mitigates the influence of outliers and initial data scale, fostering faster convergence and improved stability during training.

\begin{equation}
    \mu = \frac{1}{|x|}\sum_{x_i \in x} x_i
    \label{eq:norm_mean}
\end{equation}

\begin{equation}
    \sigma^2 = \frac{1}{|x|}\sum_{x_i \in x} (x_i -\mu)^2 = \frac{1}{|x|}\sum_{x_i \in x} x_i^2 -\mu^2
    \label{eq:norm_var}
\end{equation}

\begin{equation}
    x_{Norm} = \frac{x-\mu}{\sigma}
    \label{eq:norm}
\end{equation}

\begin{equation}
    \hat y = \sigma*\hat{y}_{Norm} + \mu =\sigma*Net(x_{Norm}) + \mu
    \label{eq:denorm}
\end{equation}

\subsection{Time Dimension Projection}
In TEFN, we employ a linear layer $project(\cdot)$ to project dimensionality to an expansion dimensionality, transforming the input time series $x$ with a length of $L_{in}$ into a sequence $x'$ with a length of $L_{in} + L_{pred}$ in \Eqref{eq:proj}. Here, $L_{pred}$ represents the desired length of the prediction horizon. This operation is analogous to the time projection mechanism utilized in other neural network architectures, such as TimesNet \cite{TimesNet}.

\begin{equation}\label{eq:proj}
    x' = project (x) = x \times W_p + b
\end{equation}

By extending the input vector to match the prediction length, the time projection layer facilitates the direct generation of future values based on the learned temporal patterns. This approach simplifies the model design and enables efficient prediction without the need for iterative or recursive computations, as commonly observed in autoregressive models.

\subsection{Basic Probability Assignment Module}
In TEFN, we leverage the principles of Evidence Theory to model the uncertainty and ambiguity inherent in time series data. According to Evidence Theory, each dimension of the time series can be represented as a mass function within a common event space. This event space is typically defined as the power set $2^S$ of the sample space $S$, where $S$ represents the possible values or labels for the data. Since time series forecasting is a regression problem, the size of the sample space needs to be used as a hyperparameter. It can be considered that the BPA operation can be regarded as an increase in dimension, and the size of the new dimension is the same as the power set. This operation expands the different representations of the original elements on the power set. The simplest form of the generating function is a linear function, which can be understood as a fuzzy membership function. Of course, it can also be a nonlinear function. In neural networks, a nonlinear layer is usually added after the linear transformation.

This representation allows us to capture the fuzzy features of the input time series, where each dimension generates a mass distribution within the event space. To achieve this, we utilize parameterized fuzzy membership functions, as defined in \Eqref{eq:bpa}. Here, $D$ represents the dimensionality of the time series, encompassing both the time dimension $T$ and the channel dimension $C$. The indices $i$ and $j$ correspond to the specific dimensions within $D$, while $k$ represents the index of the event space dimension, denoted as $F$. The fuzzy membership function $\mu_k$ is defined in \Eqref{eq:bpa} where $w_{D,j,k}$ and $b_{D,j,k}$ are the slope and intercept parameters of the membership function. The intuitive understanding of \Eqref{eq:bpa} is that the different values $j$ on the input data dimension $D$ are expanded into power sets, where $k$ corresponds to different elements in the power set. These parameters are learned during training and allow the model to capture the varying degrees of membership for each data point within the event space.

\begin{equation}
    m_{D,i,j,k} = \mu(x_{Norm,i,j}) = w_{D,j,k}*x_{Norm,i,j} + b_{D,j,k}
    \label{eq:bpa}
\end{equation}

\Figref{fig:bpa} shows the visible process of BPA module in time dimension. It should be noted that this is not matrix multiplication, but a multiplication of the same number of different parameters for each element, and the results are combined into a new dimension. The values in the \Figref{fig:bpa} corresponding to different time steps are extended to the mass of the fuzzy set memship function measure in the event space. These membership functions are shared among different channels. Therefore, the size of the newly added dimension is $2^{|S|}$, reflecting the mass function of different events. The operation on the channel dimension is symmetrical, and it shares parameters for different time steps.

By employing evidence theory and fuzzy membership functions, TEFN effectively models the uncertainty in time series data and enables the generation of robust and accurate predictions. This approach provides a principled way to handle the inherent fuzziness and ambiguity in real-world time series, leading to improved performance and reliability in various applications. Unlike the Bayesian layer and the probability layer, BPA considers imprecise probabilities, allowing for the uncertainty of the probability modeling itself, rather than using probabilities as a basis to estimate uncertainty. An intuitive example is the DSR case in Section \ref{sec:ds}.

To introduce non-linear characteristics into TEFN and enhance its modeling capabilities, we can modify the fuzzy membership function used in the BPA module. Instead of the linear form presented in \Eqref{eq:bpa}, we can adopt non-linear functions that capture more complex relationships within the data. Several options exist for non-linear membership functions, including ReLU Function, Tanh Function, Piecewise Functions, etc. By utilizing non-linear membership functions, TEFN can effectively model the non-linear dynamics and interactions within the time series data. This enhancement enables TEFN to capture more subtle patterns and relationships, leading to improved accuracy and generalization in various time series forecasting tasks.

\subsection{Expectation Fusion}
TEFN addresses the challenge of handling multivariate time series data with two distinct dimensions: the time dimension $T$ and the channel dimension $C$. This results in the paralleled generation of different mass distributions for each dimension, denoted as $m_C$ and $m_T$. To integrate these mass distributions and produce a unified prediction, TEFN employs a fusion method based on expectations, as defined in \Eqref{eq:fusion}. Here, $y_{j,k}$ represents the fusion parameters, and $m'_{i,j,k}$ corresponds to the transformed mass function for each dimension. The fusion operation in \Eqref{eq:fusion} can be interpreted as a linear transformation of the mass functions, where the summation of $y_{j,k}*m'_{i,j,k}$ effectively combines the information from different dimensions. The expectation is then calculated by summing the transformed mass functions, resulting in a single mass function representing the fused prediction.

\begin{equation}
    \begin{split}
        \hat{y}_{Norm,D,i,j} = E_{D,F}(y_{j,k}) & = \sum_{\substack{j \in [1, |D|] \\ k \in [1, |2^S|]}} y_{j,k}*m'_{i,j,k} \\
                                                & = \sum_{\substack{j \in [1, |D|] \\ k \in [1, |2^S|]}} m_{i,j,k}
    \end{split}
    \label{eq:fusion}
\end{equation}

TEFN deliberately avoids using the DSR for several reasons:
\begin{enumerate}
    \item The computational complexity of DSR is significantly higher compared to the expectation-based fusion method. Additionally, DSR involves point-to-point multiplication of mass distributions within the same dimension, which can lead to accuracy loss when dealing with high-dimensional data due to finite precision limitations.

    \item DSR is sensitive to extreme distributions, particularly those with a single element having a value of 1. This sensitivity can result in the final fused result being dominated by that single distribution, potentially leading to conflicts and loss of information from other distributions.
\end{enumerate}

By employing the expectation-based fusion method, TEFN achieves a balance between computational efficiency and accuracy. This approach effectively integrates information from different dimensions while mitigating the risks of conflicts and accuracy loss associated with DSR. The final normalized prediction results are then aggregated across dimensions using the equation in \Eqref{eq:predict_norm} and restored to their original scale through the de-normalization layer.

\begin{equation}
    \hat{y}_{norm} = \sum_{D \in \{T,C\}} \hat{y}_{Norm,D} = \hat{y}_{Norm,T} + \hat{y}_{Norm,C}
    \label{eq:predict_norm}
\end{equation}

\subsection{Comparison With Other Lightweight Forecasting Models}
Common lightweight models, such as DLinear \cite{DLinear} and LightTS \cite{LightTS}, focus on the characteristics of the time series itself through different approaches. The main differences between TEFN and other lightweight models are as follows:
\begin{itemize}
    \item \textbf{Theoretical Foundation}: While DLinear employs seasonal-trend decomposition and LightTS uses sampling techniques, evidence-based framework of TEFN fundamentally differs by modeling uncertainty through mass functions rather than direct feature extraction. This allows TEFN to handle ambiguous temporal patterns that challenge traditional decomposition methods.
    \item \textbf{Feature Engineering}: DLinear requires explicit decomposition of trends and seasonality, making it sensitive to decomposition accuracy. TEFN eliminates this need through its BPA module's automatic uncertainty quantification, as shown in \Eqref{eq:fusion}.
    \item \textbf{Computational Efficiency}: Though all models emphasize efficiency, TEFN achieves parameter reduction through expectation fusion rather than architectural simplification. Expectation fusion directly adds the results of time BPA and channel BPA, without multiplication operation that the acceleration mode is similar to \cite{ma2024era}, and without introducing new modules. The parameter amount occupied in fusion operation is $0$. Also, this will become a potential limitation, because in the engineering application scenarios of TEFN, due to the different characteristics of the data, other fusion methods may be used, such as concat and then use the linear link layer for fusion, or even use the attention mechanism with larger parameters for fusion.
\end{itemize}

\section{Experiments} \label{sec:exp}
The experimental section aims to comprehensively evaluate TEFN's performance in long-term time series forecasting. Utilizing diverse datasets, TEFN will be benchmarked against established models, measuring accuracy, efficiency, and robustness. We will analyze its performance across different prediction horizons, assess its sensitivity to hyperparameters, and conduct ablation studies to understand the impact of individual components. Lastly, we will explore TEFN's interpretability by analyzing the basic probability assignments generated by its BPA module.

\subsection{Dataset}
Our research employs the TEFN for the challenging task of long-term time series prediction. To ensure the model's effectiveness and robustness across diverse scenarios, we meticulously selected five representative and extensively used datasets. These datasets encompass a wide range of real-world applications, providing a comprehensive evaluation of TEFN's performance. The datasets are Electricity \cite{ecldata}, ETT (4 subsets) \cite{Informer}, Exchange \cite{Autoformer}, Traffic \cite{trafficdata}, and Weather \cite{weatherdata}. Each of these datasets is characterized by its large size and diverse features, offering a rich and challenging environment for evaluating TEFN's capabilities. The extensive nature of the data ensures that the model is tested on a wide range of scenarios, providing valuable insights into its performance and generalizability. More dataset features are in the \Tabref{tab:data_de}.

\begin{table}[t]

    \caption{The quantitative features of the dataset}
    \label{tab:data_de}
    \centering
    \resizebox{\linewidth}{!}{
        \begin{threeparttable}
            \begin{small}
                \renewcommand{\multirowsetup}{\centering}
                \setlength{\tabcolsep}{12pt}
                \begin{tabular}{c|c|c}
                    \toprule
                    Dataset      & Dimension & Dataset Size          \\
                    \toprule
                    Electricity  & 321       & (18317, 2633, 5261)   \\
                    \midrule
                    ETTm1, ETTm2 & 7         & (34465, 11521, 11521) \\
                    \midrule
                    ETTh1, ETTh2 & 7         & (8545, 2881, 2881)    \\
                    \midrule
                    Traffic      & 862       & (12185, 1757, 3509)   \\
                    \midrule
                    Weather      & 21        & (36792, 5271, 10540)  \\
                    \midrule
                    Exchange     & 8         & (5120, 665, 1422)     \\
                    \bottomrule
                \end{tabular}
            \end{small}
        \end{threeparttable}}
\end{table}

\begin{table*}[t]
    \caption{This table presents a comprehensive comparison of prediction errors across various models for long-term time series forecasting tasks. The models are evaluated on four different prediction lengths: $L_{pred} \in \{96, 192, 336, 720\}$, providing a thorough assessment of their performance under varying forecasting horizons. The color coding in the table highlights the ranking of each model's performance across all datasets and prediction lengths, allowing for easy identification of the top-performing models.}
    \label{tab:full_compare}
    \renewcommand{\arraystretch}{0.85}
    \centering
    \resizebox{\linewidth}{!}{
        \begin{threeparttable}
            \begin{small}
                \renewcommand{\multirowsetup}{\centering}
                \setlength{\tabcolsep}{2pt}
\begin{tabular}{c|c|cc|cc|cc|cc|cc|cc|cc|cc|cc|cc|cc|cc|cc}
                    \toprule
                    \multicolumn{2}{c}{\multirow{2}{*}{Models}}                               &
                    \multicolumn{2}{c}{\rotatebox{0}{\scalebox{0.85}{\textbf{TEFN}}}}         &
                    \multicolumn{2}{c}{\rotatebox{0}{\scalebox{0.85}{\textbf{iTransformer}}}} &
                    \multicolumn{2}{c}{\rotatebox{0}{\scalebox{0.85}{\textbf{Rlinear}}}}      &
                    \multicolumn{2}{c}{\rotatebox{0}{\scalebox{0.85}{\textbf{PatchTST}}}}     &
                    \multicolumn{2}{c}{\rotatebox{0}{\scalebox{0.85}{\textbf{Crossformer}}}}  &
                    \multicolumn{2}{c}{\rotatebox{0}{\scalebox{0.85}{\textbf{TiDE}}}}         &
                    \multicolumn{2}{c}{\rotatebox{0}{\scalebox{0.85}{\textbf{TimesNet}}}}     &
                    \multicolumn{2}{c}{\rotatebox{0}{\scalebox{0.85}{\textbf{ETSformer}}}}    &
                    \multicolumn{2}{c}{\rotatebox{0}{\scalebox{0.85}{\textbf{LightTS}}}}      &
                    \multicolumn{2}{c}{\rotatebox{0}{\scalebox{0.85}{\textbf{DLinear}}}}      &
                    \multicolumn{2}{c}{\rotatebox{0}{\scalebox{0.85}{\textbf{FEDformer}}}}    &
                    \multicolumn{2}{c}{\rotatebox{0}{\scalebox{0.85}{\textbf{Stationary}}}}   &
                    \multicolumn{2}{c}{\rotatebox{0}{\scalebox{0.85}{\textbf{Autoformer}}}}                                                                                                                                                                                                                                                                                                                                                                                                                                                                              \\
\bottomrule
                \end{tabular}
            \end{small}
        \end{threeparttable}
    }
\end{table*}

\subsection{Comparative Baseline}
To thoroughly evaluate the performance of the Time Evidence Fusion Network (TEFN), we compare it against a comprehensive set of state-of-the-art baseline models. These models represent various approaches and techniques commonly used in time series forecasting, providing a robust benchmark for assessing TEFN's capabilities. The selected baseline models are as follows: Transformer-based Models, including iTransformer (ICLR 2024) \cite{iTransformer}, Crossformer (ICLR 2023) \cite{Crossformer}, FEDformer (ICML 2022) \cite{FEDformer}, ETSformer \cite{ETSformer}, Stationary (NeurIPS 2022) \cite{Stationary}, and Autoformer (NeurIPS 2021) \cite{Autoformer}; Linear Models, such as Rlinear (ICLR 2024) \cite{Rlinear}, DLinear (AAAI 2023) \cite{DLinear}, and LightTS \cite{LightTS}; and Other Models, like TiDE (ICLR 2023) \cite{TiDE} and TimesNet (ICLR 2023).

By comparing TEFN against this diverse set of baseline models, we can comprehensively assess its performance and identify its unique advantages and contributions to the field of time series forecasting.

\subsection{Implementation Details}
Due to limited experimental equipment, our experimental environment was a regular desktop computer with a single CPU and a single GPU without paralleled computing of multi-GPU. Our experimental equipment is configured with AMD Ryzen9 7950x 4.5GHz 16 Cores, 64GB RAM DDR5 4800MHz, and NVIDIA RTX 4090 D 24GB. The proposed model TEFN is implemented in the PyTorch \cite{Pytorch}. For the sake of fairness, we used an open-source benchmark \footnote{The open-source benchmark \textit{Time-Series-Library} located at \url{https://github.com/thuml/Time-Series-Library} which contains comparison models and corresponding experiment configurations. The repository address for our model is \url{https://github.com/ztxtech/Time-Evidence-Fusion-Network}.}.
The loss function selected during the experiment was Mean Square Error (MSE), and the optimizer selected Adam \cite{Adam}. We used Mean Absolute Error (MAE) and MSE in \Eqref{eq:mae} and \Eqref{eq:mse} as error indicators in the experiment where $y_i$ is the actual value and $\hat{y}_i$ is the predicted value.

\begin{equation}
    MAE(y,\hat{y}) = \frac{1}{n}\sum_{i=1}^{n}|y_i-\hat{y}_i|
    \label{eq:mae}
\end{equation}

\begin{equation}
    MSE(y,\hat{y}) = \frac{1}{n}\sum_{i=1}^{n}(y_i-\hat{y}_i)^2
    \label{eq:mse}
\end{equation}

\subsection{Forecasting Result}
In the comparative experiment, we conducted a thorough comparison of 5 datasets and 12 comparative models in terms of forecasting length $L_{pred} \in \{96,192,336,720\}$ in \Tabref{tab:full_compare}. The results showcase the remarkable capabilities of the Time Evidence Fusion Network (TEFN) in long-term time series forecasting. TEFN consistently achieves state-of-the-art (SOTA) performance across various datasets and prediction horizons, highlighting its effectiveness and versatility in capturing complex temporal patterns. Notably, TEFN achieves this with a significantly smaller model size compared to models such as iTransformer \cite{iTransformer} and PatchTST \cite{PatchTST}, demonstrating its efficiency and scalability.

TEFN has significant advantages over other models in terms of prediction accuracy, as evidenced by the MSE and MAE metrics. Specifically, in terms of MSE, TEFN is $0.028$ smaller than iTransformer \cite{iTransformer} and $0.025$ smaller than PatchTST \cite{PatchTST} in the task of predicting a length of $L_{pred} = 720$ on the ETTh1 dataset \cite{Informer}. Similarly, in terms of MAE, TEFN once again showed a clear lead over iTransformer. TEFN was $0.008$ smaller than iTransformer \cite{iTransformer} and $0.007$ smaller than PatchTST \cite{PatchTST} in predicting a length of $L_{pred}=96$ on the Exchange dataset \cite{Autoformer}. These results highlight the effectiveness and robustness of TEFN in time series prediction, demonstrating its ability to provide more accurate predictions compared to contemporary methods.

\begin{figure*}[htbp]
    \centering
    \subfigure[\textbf{TEFN (Ours)-Electricity} \label{fig:ECL-s}]{\includegraphics[width=0.2\linewidth]{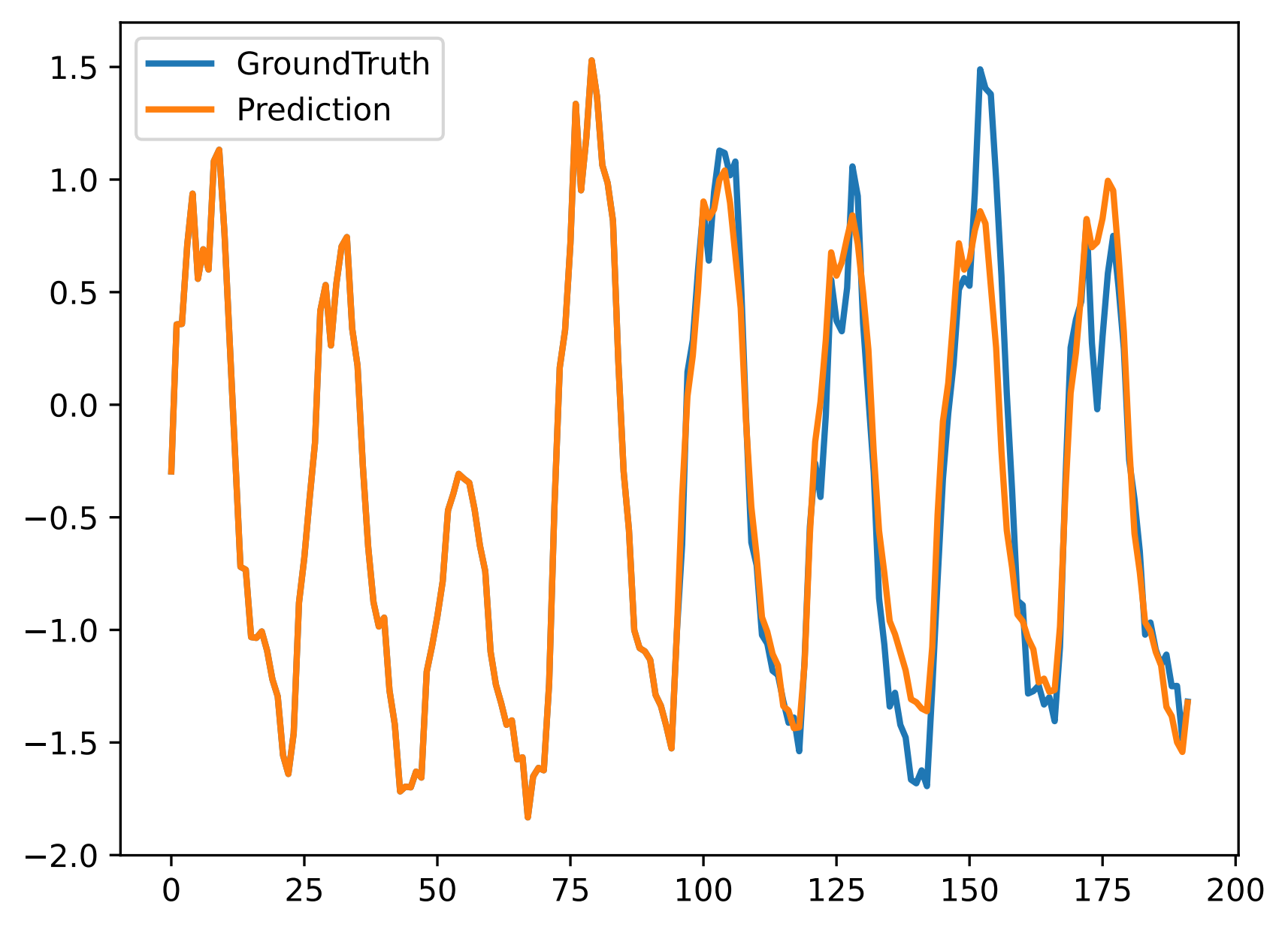}}
    \subfigure[PatchTST-Electricity]{\includegraphics[width=0.2\linewidth]{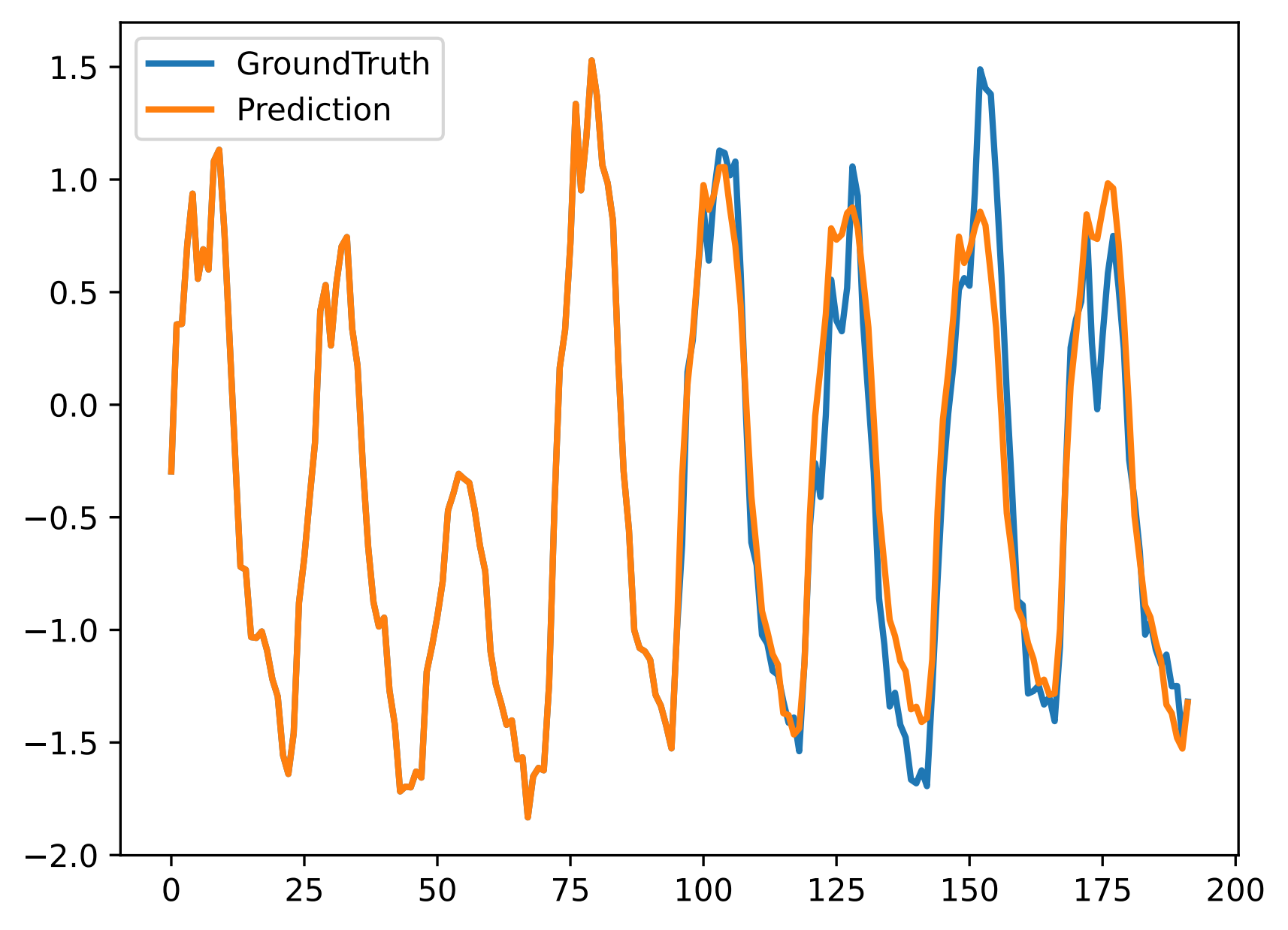}}
    \subfigure[Crossformer-Electricity]{\includegraphics[width=0.2\linewidth]{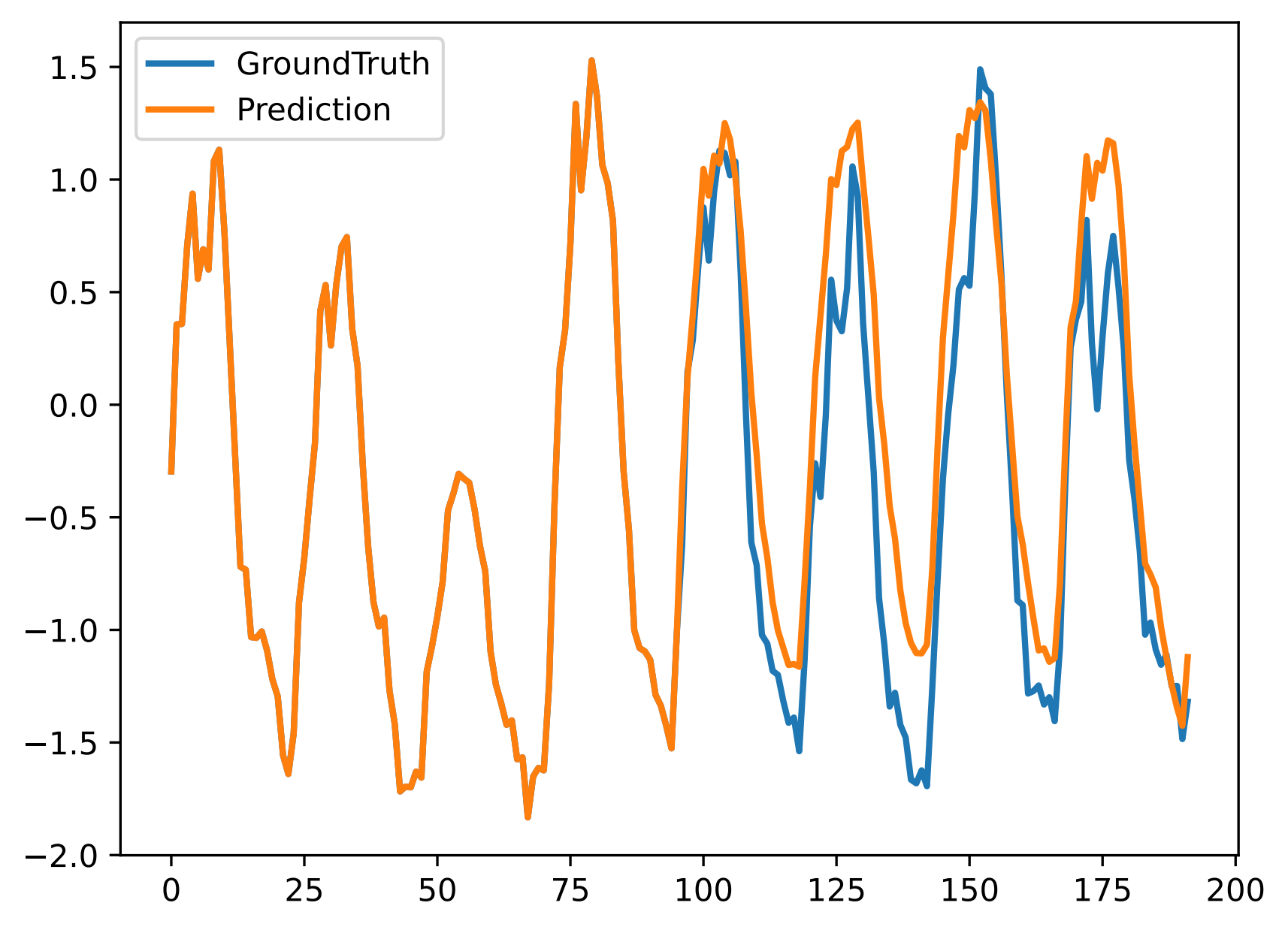}}
    \subfigure[TimesNet-Electricity]{\includegraphics[width=0.2\linewidth]{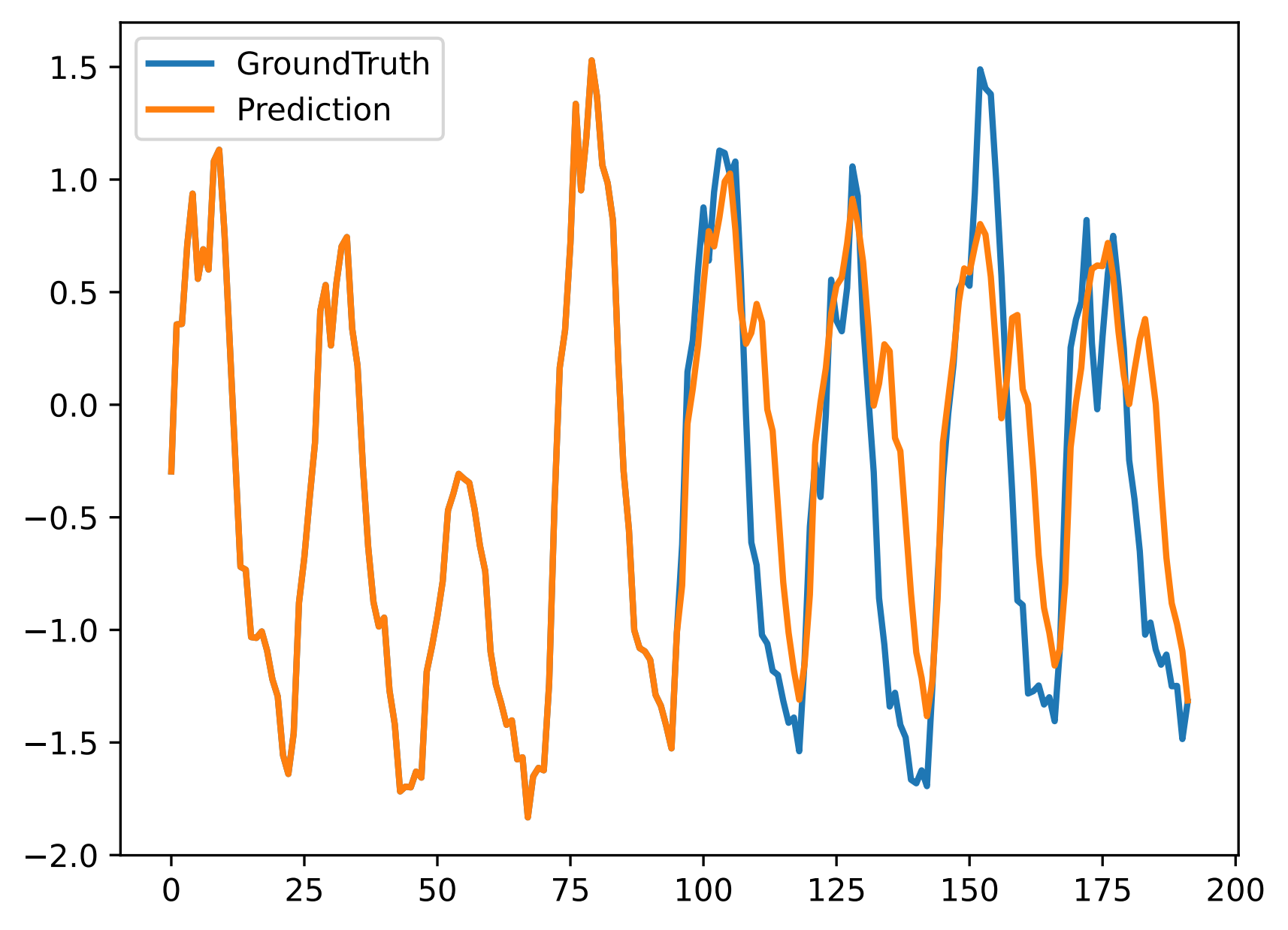}}
    \\
    \subfigure[DLinear-Electricity]{\includegraphics[width=0.2\linewidth]{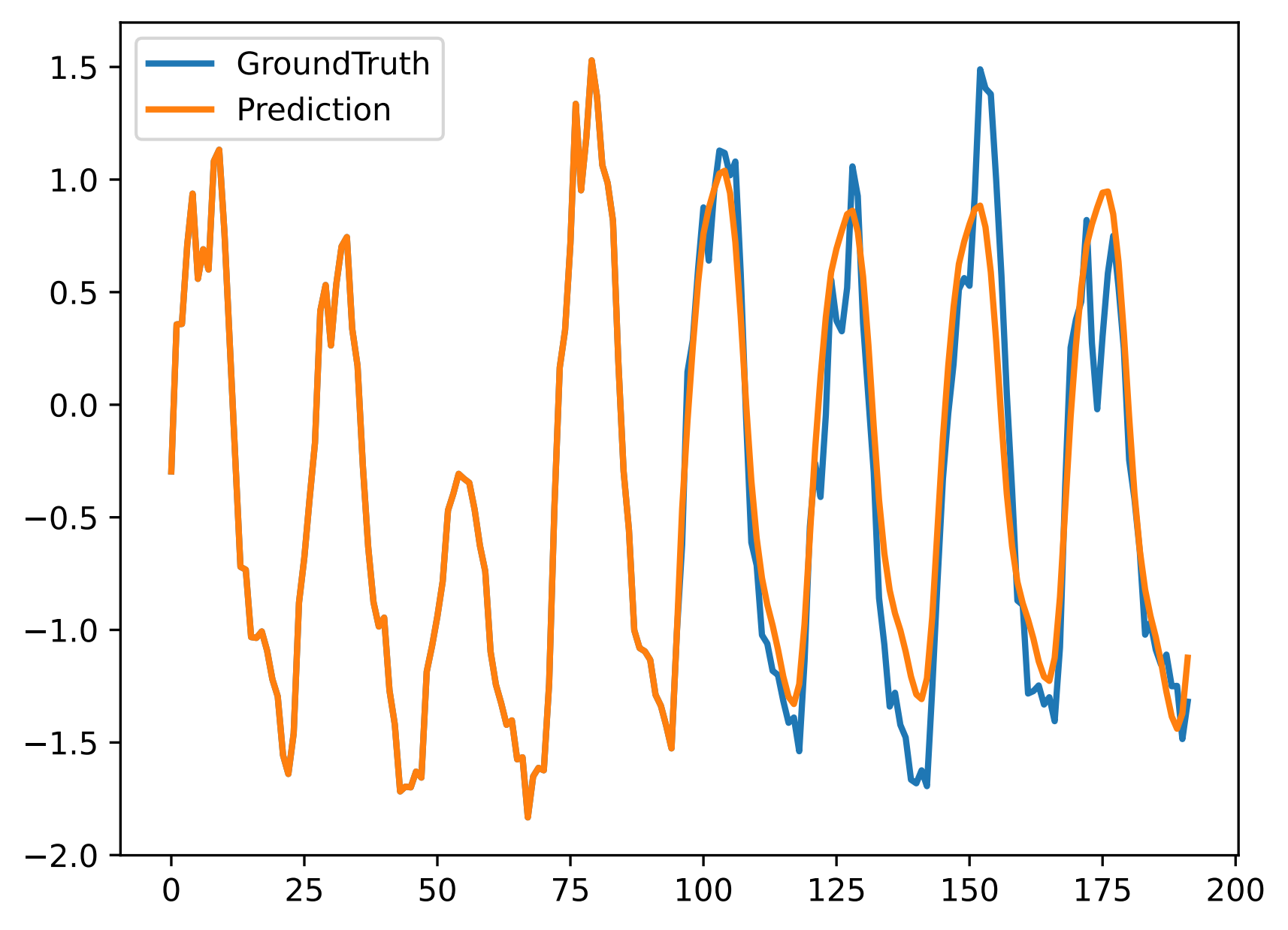}}
    \subfigure[Stationary-Electricity \label{fig:ECL-e}]{\includegraphics[width=0.2\linewidth]{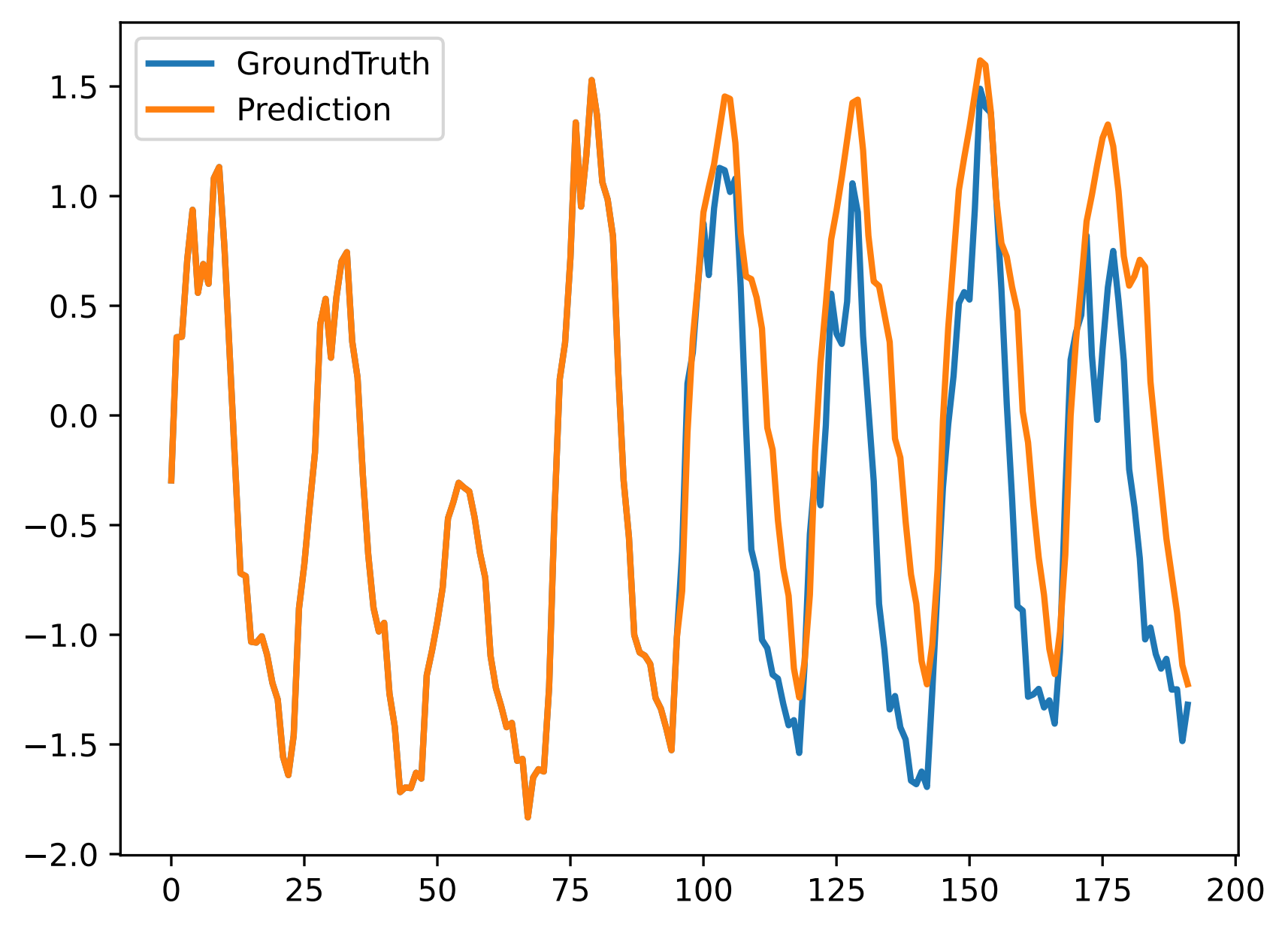}}
    \subfigure[\textbf{TEFN (Ours)-ETTm1} \label{fig:ETTm1-s}]{\includegraphics[width=0.2\linewidth]{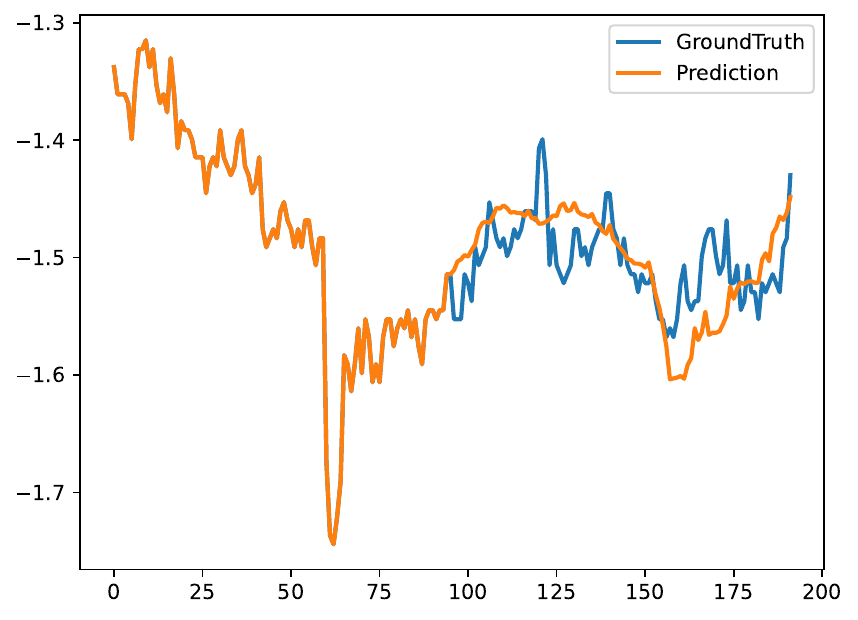}}
    \subfigure[PatchTST-ETTm1]{\includegraphics[width=0.2\linewidth]{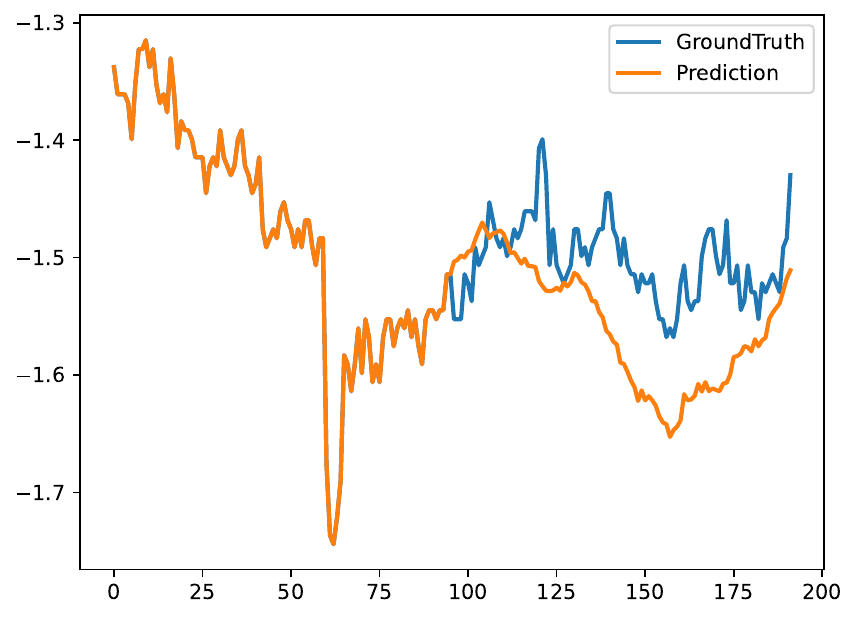}}
    \\
    \subfigure[Crossformer-ETTm1]{\includegraphics[width=0.2\linewidth]{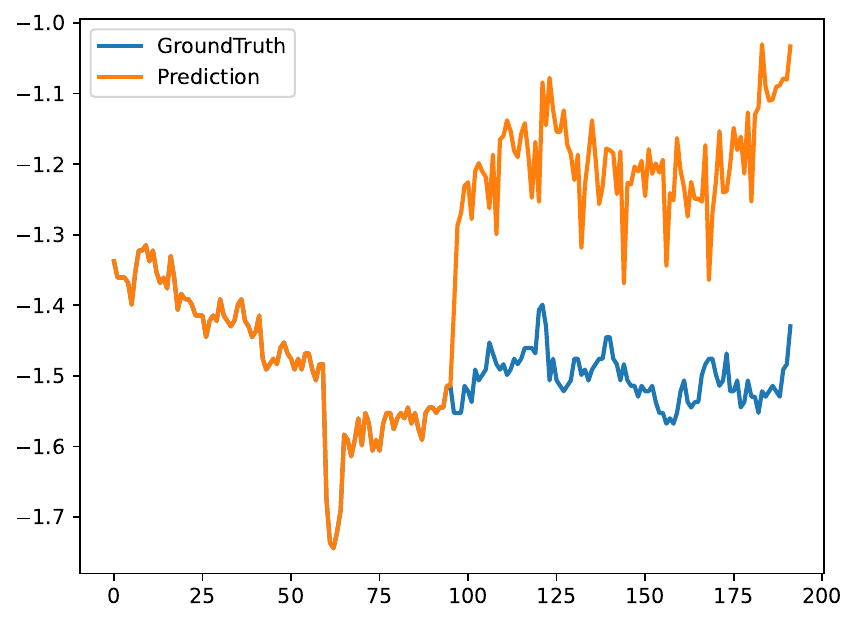}}
    \subfigure[TimesNet-ETTm1]{\includegraphics[width=0.2\linewidth]{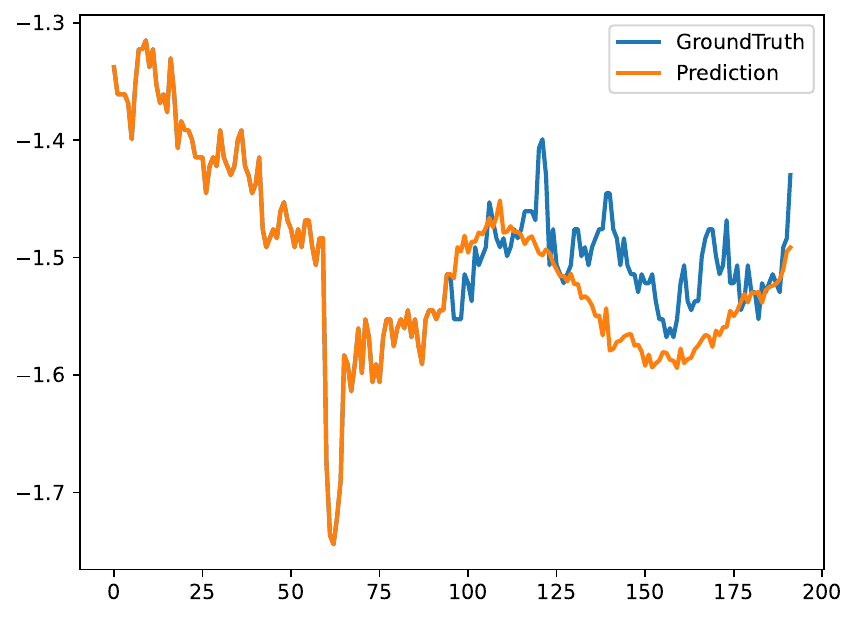}}
    \subfigure[DLinear-ETTm1]{\includegraphics[width=0.2\linewidth]{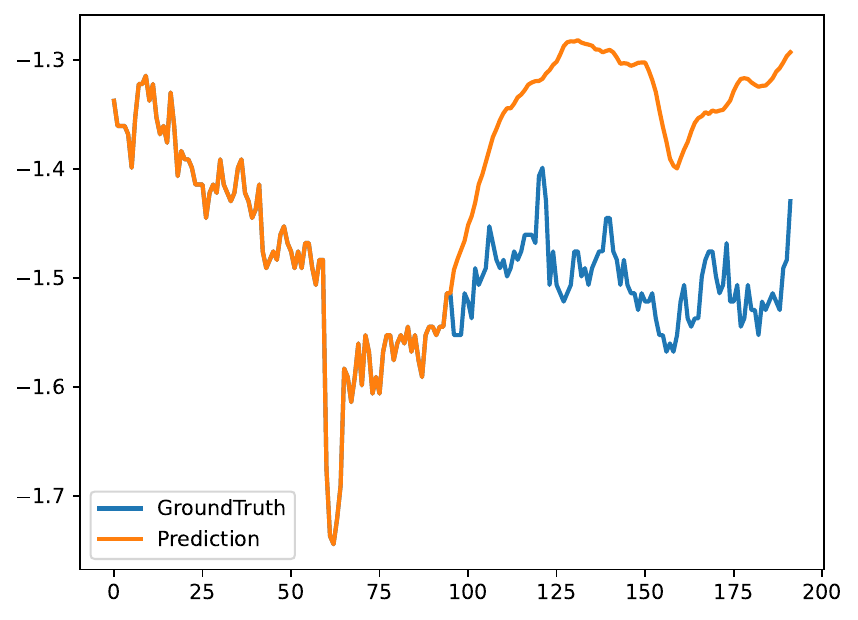}}
    \subfigure[Stationary-ETTm1 \label{fig:ETTm1-e}]{\includegraphics[width=0.2\linewidth]{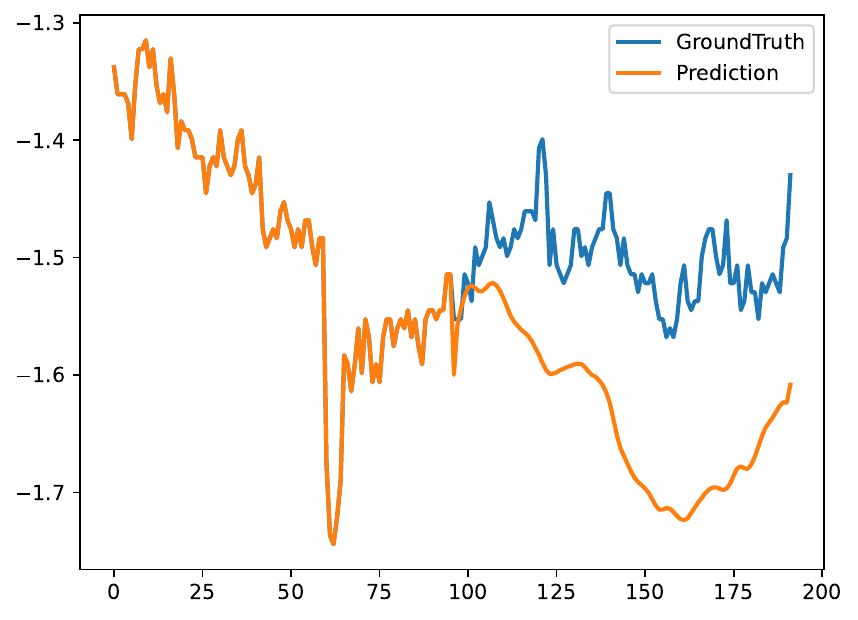}}
    \caption{Visualization of long-term forecasting tasks: \Figref{fig:ECL-s} - \Figref{fig:ECL-e} in Electricity dataset with $L_{pred}=96$ and \Figref{fig:ETTm1-s} - \Figref{fig:ETTm1-e} in ETTm1 dataset with $L_{pred}=96$}
    \label{fig:vis_ecl_96}
\end{figure*}

We visualized the predictions of TEFN and comparative models to better demonstrate the predictive performance of different models. We take the task of Electricity-96 \cite{ecldata} and ETTm1-96 as an example and selected typical models PatchTST \cite{PatchTST}, Crossformer \cite{Crossformer}, TimesNet \cite{TimesNet}, DLinear \cite{DLinear} and Stationary \cite{Stationary} for visualization in \Figref{fig:vis_ecl_96}. From the visualization results, TEFN fits the details and trends of time series better.

\subsection{Efficiency Comparison}\label{sec:eff}
\begin{figure}[htbp]
    \centering
    \includegraphics[width=0.9\linewidth]{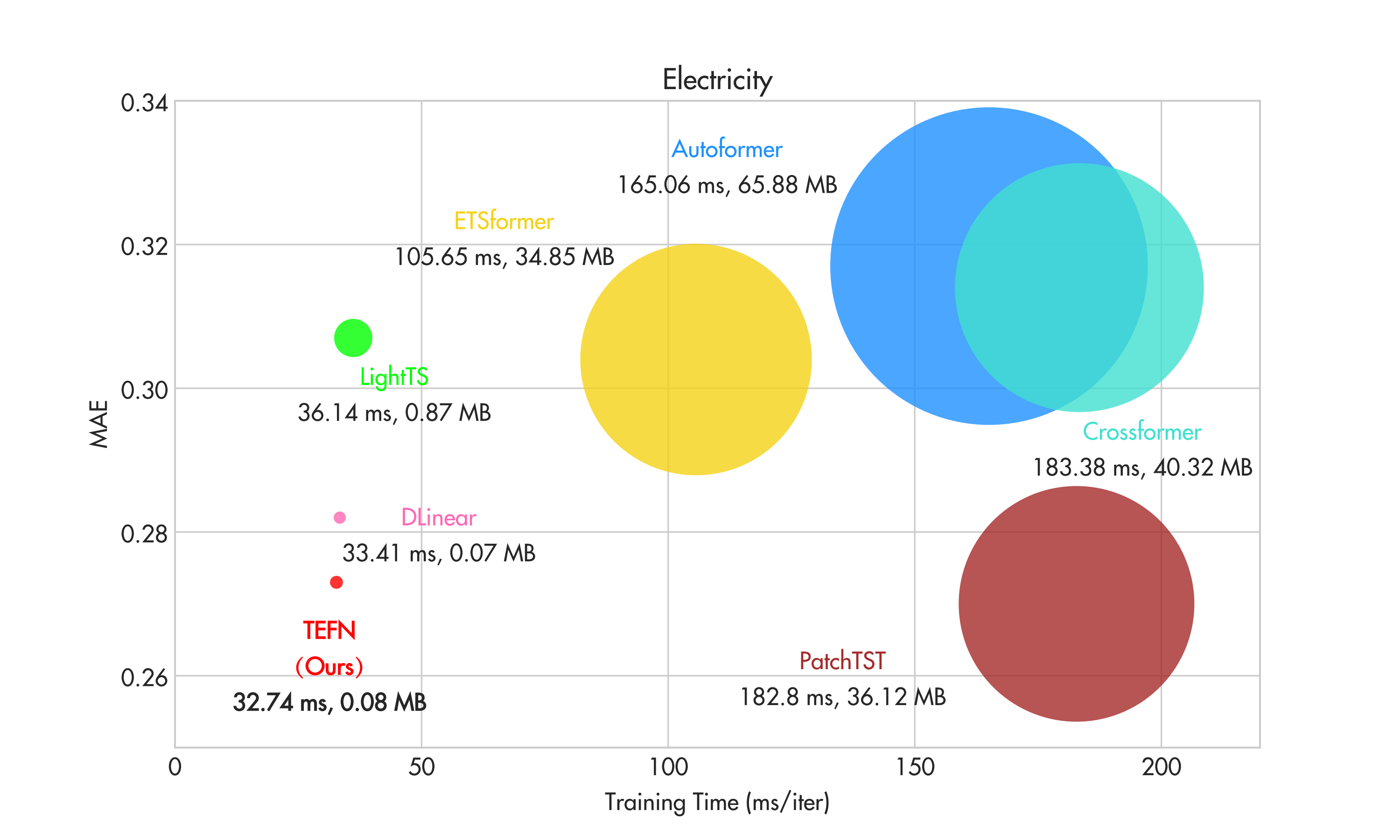}
    \caption{Model efficiency comparison. Results are from the prediction task Electricity-96, where the model size refers to the size of the binary model files, and iteration time is averaged each iteration. The models beyond the canvas have not been drawn.}
    \label{fig:size}
\end{figure}

TEFN is a compact model compared to the parameters of the Transformer. In order to effectively compare the efficiency of the model, we chose the dataset Electricity with a large dataset for performance comparison. In the task Electricity-96, we measured the average time, MSE, and exported model file size for iterating over one sample. Different models occupy varying amounts of GPU and memory. Therefore, we used the size of the saved binary model files as the model size. We visualized this with a bubble chart in \Figref{fig:size} . It can be seen that TEFN achieves a low prediction error with an extremely small number of parameters. Meanwhile, it has the shortest average time for iterating over one sample, demonstrating its high efficiency. TEFN maintains almost DLinear \cite{DLinear} performance and almost PatchTST \cite{PatchTST} accuracy, making it a performance error balanced model.

It is particularly relevant to note that TEFN has demonstrated strong performance on portable devices. As evidenced by the code repository, TEFN's training and prediction processes have been successfully executed on a Macbook Air 2020 equipped with Apple Silicon M1 and 8 GB RAM. This finding suggests that TEFN's computational efficiency is well-suited for resource-constrained environments, contrasting with the challenges typically encountered by Transformer-based models such as Crossformer \cite{Crossformer} when deployed on portable computers.

According to the process of BPA, the time complexity of BPA is $O(n*2^{|S|})$ where $n$ is input size and $|S|$ is sample space size. For the TEFN model, the specific time complexity is $O(C*L*2^{|S|})$ where $C$ is channel size, and $L$ is the length of time series. $C$ and $|S|$ are constants and satisfy $C, |S| << L$. Hence, the final time complexity is $O(L)$. So TEFN can maintain positive performance in long-term forecasting tasks. Another factor affecting experimental efficiency is whether TEFN is linear or nonlinear. We chose to use a linear BPA module in the experimental section because of its high efficiency. Nonlinear modules also exist, and only the membership functions need to be replaced. BPA needs to correspond to different types of nonlinear time series and design membership functions. In the future, we will explore adaptive nonlinear functions. The fitting of nonlinear functions also brings higher computational costs, therefore, in the case of efficiency priority, the linear version of TEFN is the preferred choice.

\subsection{Hyperparameter Sensitivity}\label{sec:hs}

\begin{figure}[h]
    \centering
    \subfigure[$lr$ - $|S|$ -MSE]{\includegraphics[width=0.8\linewidth]{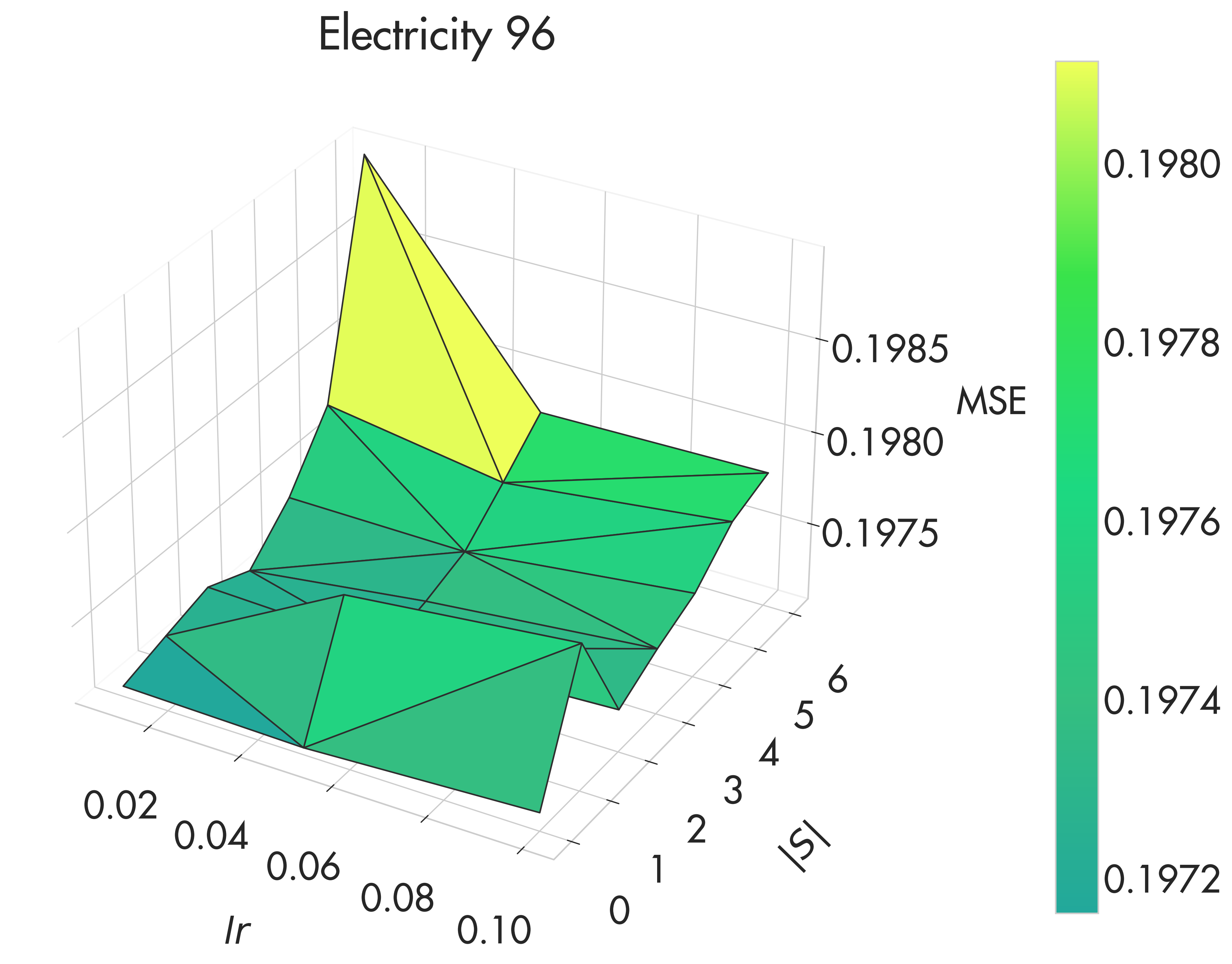}}
    \\
    \subfigure[Violin diagram of $|S|$ and MSE]{\includegraphics[width=\linewidth]{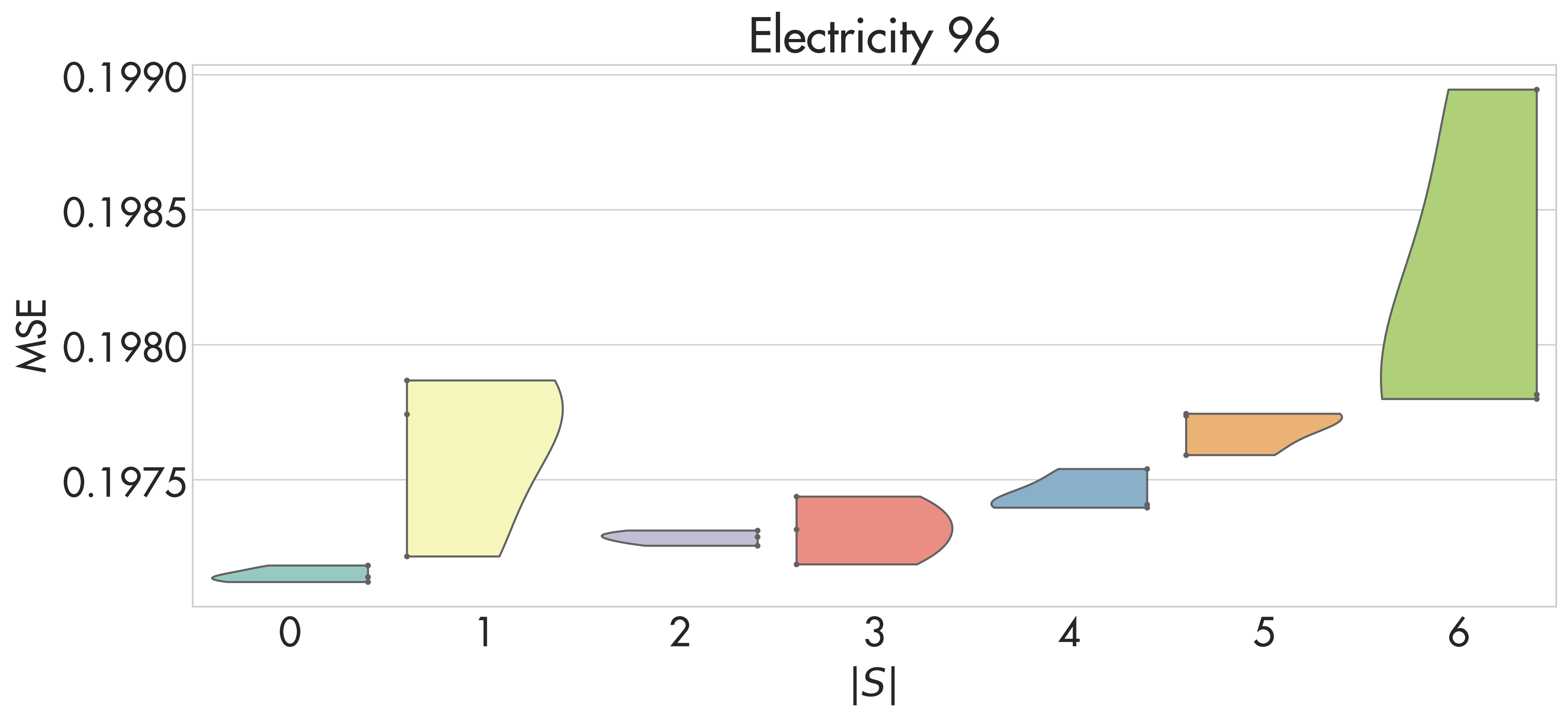}}
    \caption{Traversing error of Electricity 96 task: 3D visualization and violin diagram. The points on the number axis of the violin chart reflect the density of learning rate.}
    \label{fig:model_sen}
\end{figure}

To test the hyperparameter sensitivity of TEFN, we conducted a 2D traversal of the learning rate $lr \in \{10^{-2}, 5*10^{-2}, 10^{-1}\}$ and sample space size $|S|\in \{0,1,2,3,4,5,6\}$. We take the Electricity-96 task as an example. TEFN demonstrates weak sensitivity to parameters, but as the sample space increases, the error magnitude also increases, leading to a certain degree of overfitting in the model as shown in \Figref{fig:model_sen}. It is worth noting that the order of magnitude of the MSE in \Figref{fig:model_sen} is $10^{-4}$, while the order of magnitude of the compared MSE is $10^{-3}$, so the fluctuation of different hyperparameter in error metrics is small.

\begin{table}[t]
    \caption{Non-linearity of different datasets}
    \label{tab:nonlinear}
    \centering
    \resizebox{\columnwidth}{!}{
        \begin{threeparttable}
            \begin{small}
                \renewcommand{\multirowsetup}{\centering}
                \setlength{\tabcolsep}{2pt}
                \begin{tabular}{c|c|c|c|c|c|c|c|c|c}
                    \toprule
                    \multicolumn{2}{c}{\multirow{2}{*}{Non-linearity}} & \multicolumn{8}{c}{Dataset}                                                                        \\
                    \cmidrule(lr){3-10}
                    \multicolumn{2}{c}{}                               & Electricity                 & ETTh1 & ETTh2 & ETTm1 & ETTm2 & Exchange & Traffic & Weather         \\
                    \midrule
                    \multirow{4}{*}{MSE}                               & 96                          & 0.196 & 0.399 & 0.342 & 0.345 & 0.197    & 0.084   & 0.653   & 0.198 \\
                                                                       & 192                         & 0.195 & 0.442 & 0.452 & 0.382 & 0.275    & 0.172   & 0.602   & 0.239 \\
                                                                       & 336                         & 0.208 & 0.492 & 0.547 & 0.415 & 0.374    & 0.308   & 0.609   & 0.281 \\
                                                                       & 720                         & 0.244 & 0.516 & 0.763 & 0.484 & 0.451    & 0.792   & 0.649   & 0.346 \\
                    \midrule
                    \multirow{4}{*}{MAE}                               & 96                          & 0.279 & 0.413 & 0.393 & 0.372 & 0.291    & 0.211   & 0.401   & 0.259 \\
                                                                       & 192                         & 0.282 & 0.435 & 0.455 & 0.391 & 0.352    & 0.308   & 0.375   & 0.302 \\
                                                                       & 336                         & 0.298 & 0.467 & 0.516 & 0.417 & 0.420    & 0.413   & 0.378   & 0.331 \\
                                                                       & 720                         & 0.331 & 0.511 & 0.630 & 0.465 & 0.458    & 0.669   & 0.399   & 0.384 \\
                    \bottomrule
                \end{tabular}
            \end{small}
        \end{threeparttable}}
\end{table}

\begin{table}[t]
    \caption{The variance of model error for traversal learning rate and sample space}\label{tab:var}
    \centering
    \resizebox{\columnwidth}{!}{
        \begin{threeparttable}
            \begin{small}
                \renewcommand{\multirowsetup}{\centering}
                \setlength{\tabcolsep}{2pt}
                \begin{tabular}{c|c|c|c|c|c|c|c|c|c}
                    \toprule
                    \multicolumn{2}{c}{\multirow{2}{*}{Metric}} & \multicolumn{8}{c}{Dataset}                                                                             \\
                    \cmidrule(lr){3-10}
                    \multicolumn{2}{c}{($Var$)}                 & Electricity                 & ETTh1  & ETTh2  & ETTm1  & ETTm2  & Exchange & Traffic & Weather          \\
                    \midrule
                    \multirow{4}{*}{MSE}                        & 96                          & 0.0000 & 0.0000 & 0.0002 & 0.0002 & 0.0001   & 0.0000  & 0.0000  & 0.0001 \\
                                                                & 192                         & 0.0001 & 0.0000 & 0.0003 & 0.0004 & 0.0000   & 0.0000  & 0.0000  & 0.0002 \\
                                                                & 336                         & 0.0001 & 0.0000 & 0.0002 & 0.0004 & 0.0000   & 0.0023  & 0.0000  & 0.0001 \\
                                                                & 720                         & 0.0000 & 0.0000 & 0.0004 & 0.0001 & 0.0000   & 0.0038  & 0.0000  & 0.0001 \\
                    \midrule
                    \multirow{4}{*}{MAE}                        & 96                          & 0.0000 & 0.0000 & 0.0001 & 0.0001 & 0.0001   & 0.0000  & 0.0000  & 0.0002 \\
                                                                & 192                         & 0.0001 & 0.0000 & 0.0001 & 0.0001 & 0.0000   & 0.0000  & 0.0000  & 0.0002 \\
                                                                & 336                         & 0.0001 & 0.0000 & 0.0001 & 0.0001 & 0.0000   & 0.0009  & 0.0000  & 0.0001 \\
                                                                & 720                         & 0.0000 & 0.0000 & 0.0001 & 0.0000 & 0.0000   & 0.0007  & 0.0000  & 0.0000 \\
                    \bottomrule
                \end{tabular}
            \end{small}
        \end{threeparttable}}
\end{table}

\begin{figure}[h]
    \centering
    \includegraphics[width=0.7\linewidth]{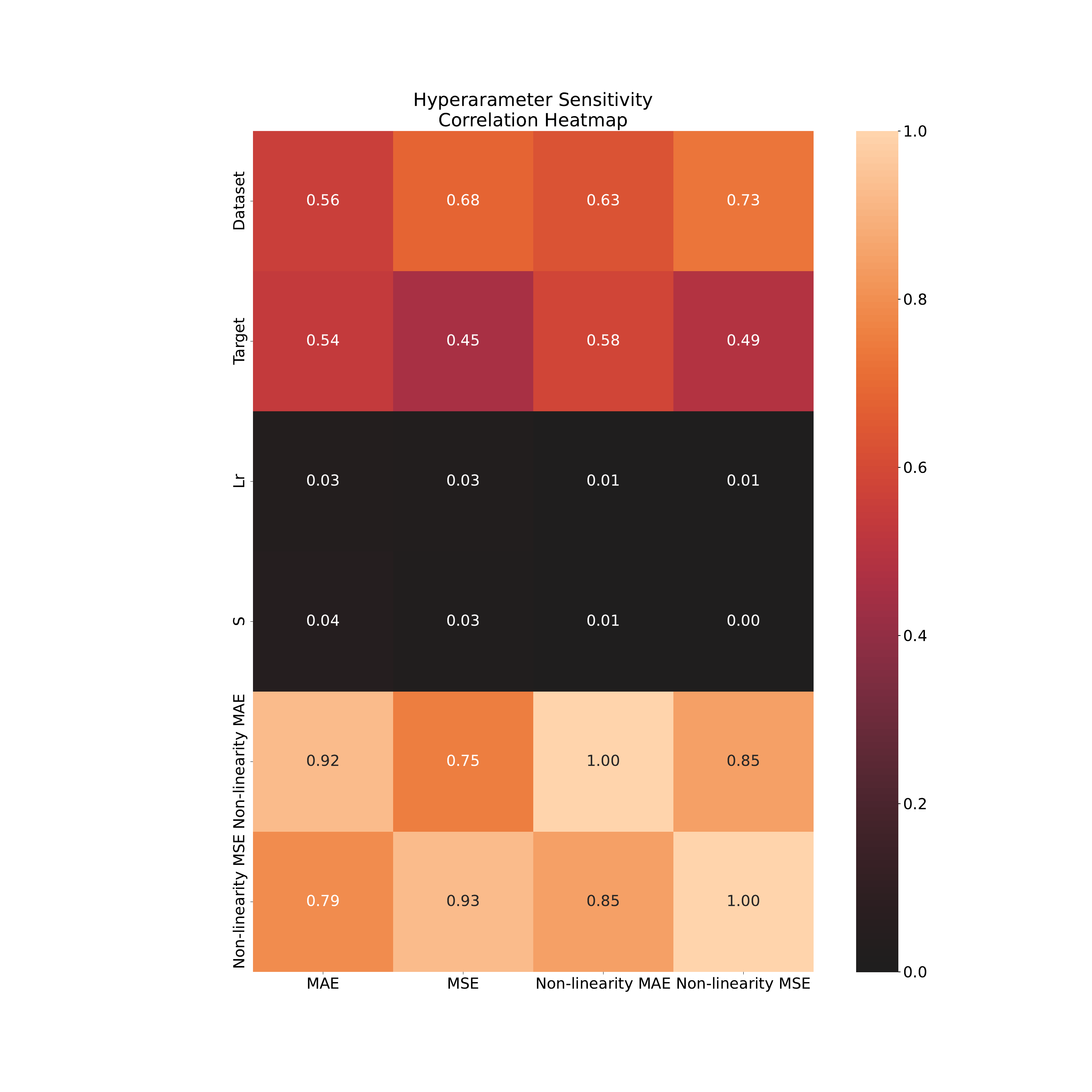}
    \caption{Heatmap of hyperparameter sensitivity correlation}
    \label{fig:hsch}
\end{figure}

Moreover, to observe the relationship between the non-linearity and the performance of TEFN, we quantified the non-linearity of different datasets in an intuitive way: We directly use a linear layer to predict all the datasets. Since all the datasets have undergone the same standardization before prediction (following the settings of the original Benchmark), the errors are at the same level. Moreover, the larger the error, the greater the non-linearity. The nonlinear experimental results are in Table \ref{tab:nonlinear}. It can be found that the longer the prediction length, the greater the degree of nonlinearity. For different datasets, the error in the Traffic dataset is relatively large at different step lengths, showing a significant non-linearity. The non-linearity of the ETT dataset is relatively stable. In addition, Exchange dataset is greatly influenced by the prediction step length, reflecting that Exchange dataset shows linearity in the short term, but as the prediction step length increases, the linearity gradually decreases. In order to more intuitively reflect the hyperparameter sensitivity of TEFN, correlation analysis \cite{leuthold1975use, Zychlinski_dython_2018} \footnote{The implementation of relevant calculations for discrete indicators such as Dataset comes from dython Package (\url{https://shakedzy.xyz/dython/}).} is conducted on non-linearity, prediction length (Target), dataset, sample space (S), Learning rate (Lr), MAE and MSE in \Figref{fig:hsch}. According to the correlation between different indicators, it can be found that TEFN is stable to the learning rate and sample space size which are almost uncorrelated with the two error indicators MSE and MAE. In terms of nonlinearity, the forecasting error of the model and the non-linearity show a high positive correlation, such as the correlation between non-linearity MSE and MSE. This also means that TEFN will have a lower prediction error on datasets with low non-linearity, while the error will be relatively higher for datasets with high non-linearity. \emph{This comparison is only limited to the cross-dataset comparison of TEFN, and cannot reflect the performance comparison between different models.} Building on this, a broader analysis offers insights into the performance of linear versus non-linear models. The observed positive correlation between non-linearity and prediction error appears to be a general phenomenon for linear models, as seen by comparing models like DLinear and RLinear in \Tabref{tab:full_compare} with the non-linearity scores in \Tabref{tab:nonlinear}. While non-linear models can potentially mitigate this issue, their success is not guaranteed and often hinges on the alignment between their architectural design and the data's specific characteristics. For instance, a model with a strong inductive bias for crossed variable relationships, such as Crossformer, may excel on one dataset such as Weather but not universally outperform simpler models on others.

This highlights a crucial tradeoff. Non-linear models offer the potential for higher accuracy on complex data but at the cost of increased computational complexity, memory usage, and potentially lower robustness. Conversely, linear models like TEFN, while having a performance ceiling on highly non-linear data, provide significant benefits in terms of efficiency in \Figref{fig:size}. In many scenarios, as evidenced by our results, they provide a highly competitive balance between performance and cost, making them a practical and effective choice for many real-world applications.

Additionally, we conducted statistical analysis on the error variance of different tasks as shown in \Tabref{tab:var}. The order of magnitude of variance is $10^{-4}$, indicating extremely small fluctuation. On the other hand, we can observe that as the prediction step length increases, the variance also tends to increase, indicating that although non-linearity can improve the parameter sensitivity of the model, due to the fewer parameters of the linear model, its amplitude is limited. Random parameter selection within the traversal range of TEFN yields stable prediction results.

\subsection{Ablation Study}\label{sec:as}
\begin{table*}[h]
    \renewcommand{\arraystretch}{1.2}
    \caption{The results of the ablation experiment}\label{tab:full_abla}
    \centering
    \resizebox{\linewidth}{!}{
        \begin{threeparttable}
            \begin{small}
                \renewcommand{\multirowsetup}{\centering}
                \setlength{\tabcolsep}{1.5pt}
\begin{tabular}{c|c|cc|cc|cc|cc|cc|cc|cc|cc|cc|cc|cc|cc|cc|cc|cc}
                    \toprule
                    \multicolumn{2}{c}{Models}                                    & \multicolumn{2}{c}{\textbf{TEFN}} & \multicolumn{4}{c}{\textbf{\textit{w/o} Norm}} & \multicolumn{4}{c}{\textbf{\textit{w/o} T}} & \multicolumn{4}{c}{\textbf{\textit{w/o} C}} & \multicolumn{4}{c}{\textbf{Prob}} & \multicolumn{4}{c}{\textbf{Concat}} & \multicolumn{4}{c}{\textbf{Relu}} & \multicolumn{4}{c}{\textbf{Tanh}}                                                                                                                                                                                                                                                                                                                                                                                                                                   \\
\bottomrule
                \end{tabular}
            \end{small}
        \end{threeparttable}}
\end{table*}
We conducted experiments on the components of TEFN, the replacement of the probability layer, the concat fusion method, and different BPA nonlinear implementations in \Tabref{tab:full_abla}. At the same time, we calculated the error change ratio $\gamma$ compared with the original TEFN in Equation \eqref{eq:gamma}.

\begin{equation}
    \label{eq:gamma}
    \gamma = \frac{\text{Metric}_{TEFN}-\text{Metric}}{\text{Metric}}*100\%
\end{equation}

\subsubsection{Components}
TEFN consists of a norm layer, a time dimension projection layer \text{Norm}, a time fusion layer \textbf{T}, and a channel fusion layer \textbf{C}. Across most forecasting tasks, the fusion $C+T$ error is lower than that of individual time $T$ or channel $C$ error. Although partial use of a single channel may further reduce prediction error, the error reduction is not significant (the error of Electricity only varies $0.001$). It displays that TEFN effectively filters out time $T$ and channel $C$ information, and individual time $T$ and channel $C$ information alone cannot effectively construct the evolution pattern of time series.
There are a few cases where the effect is not as good as the individual dimensions, but the difference is small and can be ignored. 
The forecasting error of TEFN $l_{C+T}$ will be less than or equal to using separate time $l_T$ or channel $l_C$ in \Eqref{eq:loss}.

\begin{equation}
    l_{C+T} \leqslant min(l_C,l_T)
    \label{eq:loss}
\end{equation}

Furthermore, we empirically demonstrated that removing the normalization layer led to a substantial performance improvement on certain datasets, such as the Exchange dataset. The normalization layer typically accelerates the model's training process; however, it also presents issues. During denormalization, it relies solely on the standard deviation and mean of the input samples, overlooking potential changes in these statistics. Conversely, the normalization layer proves beneficial for the ETT dataset. Without it, the model exhibits severe underfitting, as evidenced by a twofold increase in the MSE for ETTh2-720.

\subsubsection{Probability Layer}
To elucidate the disparity between the Basic Probability Assignment (BPA) and a traditional probability model, we conducted an experiment by removing two BPA layers. As depicted in Equation \eqref{eq:fusion}, when the mass function degenerates into a probability, the BPA module simplifies to an ordinary linear layer. At this juncture, the time dimension projection accomplishes an expectation prediction from a probabilistic perspective. Moreover, BPA can be regarded as an imprecise estimate of probability, as it takes into account a broader spectrum of possibilities. Consequently, the output of the BPA module increases the dimensionality of the data. Empirical results indicate that, in comparison to the original TEFN, most of the forecasting errors have risen.

\subsubsection{Concat Fusion Method}
The integration approach adopted by TEFN is expectation fusion. Therefore, we can directly add the expanded dimensions of the two BPAs to complete the prediction without introducing additional fusion layers. Here, we use the concat method for comparison. We concatenate the results of the time BPA and the channel BPA, and then use a linear layer for the final output. Experimental data shows that the concat method does not significantly improve the prediction performance. However, due to the concat operation and the addition of the fusion layer, new parameters are introduced, and more memory is occupied.

\subsubsection{Nonlinear Implementation of BPA}
We added a nonlinear activation layer after the BPA module to enhance the nonlinear fitting ability of TEFN using the residual method described in Equation \eqref{eq:res_non}. The ReLU function, commonly used in classification tasks, and the Tanh activation function, commonly used in time series analysis, have been selected here. Experimental data indicates that the performance of ReLU is not ideal because ReLU essentially filters out weights less than 0, rather than performing a continuous transformation. In contrast, Tanh improves the forecasting performance of TEFN in most scenarios because it is a smooth function suitable for time series prediction.

\begin{equation}
    \label{eq:res_non}
    x \leftarrow x + \text{act}(x)
\end{equation}

\subsection{Robustness Experiments}
TEFN is a highly robust neural network. To verify the robustness of TEFN in real prediction, a comparative experiment was conducted by adding noise to the dataset. The unit noise $\varepsilon$ follows the Gaussian distribution $N(0, 1)$ with an expectation of $0$ and a variance of $1$. Assume that the input time series is $x$, with time and channel dimensions. To adapt to the scale of the data, the transformation as Equation \eqref{eq:noise_trans} is applied to input data where $i$ is the index of channel dimension, $i$ is the index of time dimension, $x_i$ is a vector containing all time steps of channel $i$ and $std(\cdot)$ is Standard deviation function. So, the noisy data experiments are compared with the original data experiments.

\begin{equation}
    \label{eq:noise_trans}
    x'_{ij} = x_{ij} + \text{std}(x_i)*\varepsilon
\end{equation}

The experimental results of robustness testing are shown in the \Tabref{tab:rr}. Continuing from the analysis, the experimental outcomes underscore the robust nature of TEFN, where the introduction of Gaussian noise to the dataset led to only marginal increases in prediction errors, with most variations being insignificant relative to the original errors. Notably, in long-term forecasting scenarios, a counterintuitive improvement in accuracy was observed in some instances, indicating that TEFN's adaptability and learning capacity in complex, dynamic environments extend beyond mere resilience to noise, potentially leveraging subtle perturbations to enhance prediction performance.

\begin{table}[h]
    \caption{The results of robustness experiments}\label{tab:rr}
    \centering
    \resizebox{\columnwidth}{!}{
        \begin{threeparttable}
            \begin{small}
                \renewcommand{\multirowsetup}{\centering}
                \setlength{\tabcolsep}{12pt}
                \begin{tabular}{c|c|cc|cc}
                    \toprule
                    \multicolumn{1}{c}{\multirow{2}{*}{Dataset}} & \multicolumn{1}{c}{\multirow{2}{*}{Pred}} & \multicolumn{2}{c}{Original} & \multicolumn{2}{c}{Noisy}                                                                                 \\
                    \cmidrule{3-6}
                    \multicolumn{1}{c}{}                         & \multicolumn{1}{c}{}                      & MSE                          & MAE                       & MSE                                   & MAE                                   \\
                    \midrule
                    \multirow{4}{*}{ETTm1}
                                                                 & 96                                        & 0.343                        & 0.367                     & 0.414 \firstres{($\uparrow$ 0.071)}   & 0.411 \firstres{($\uparrow$ 0.044)}   \\
                                                                 & 192                                       & 0.381                        & 0.383                     & 0.443 \firstres{($\uparrow$ 0.062)}   & 0.429 \firstres{($\uparrow$ 0.046)}   \\
                                                                 & 336                                       & 0.414                        & 0.404                     & 0.499 \firstres{($\uparrow$ 0.085)}   & 0.458 \firstres{($\uparrow$ 0.054)}   \\
                                                                 & 720                                       & 0.475                        & 0.438                     & 0.381 \thirdres{($\downarrow$ 0.094)} & 0.398 \thirdres{($\downarrow$ 0.04)}  \\
                    \midrule
                    \multirow{4}{*}{ETTm2}
                                                                 & 96                                        & 0.181                        & 0.264                     & 0.254 \firstres{($\uparrow$ 0.073)}   & 0.312 \firstres{($\uparrow$ 0.048)}   \\
                                                                 & 192                                       & 0.246                        & 0.304                     & 0.314 \firstres{($\uparrow$ 0.068)}   & 0.350 \firstres{($\uparrow$ 0.046)}   \\
                                                                 & 336                                       & 0.307                        & 0.343                     & 0.411 \firstres{($\uparrow$ 0.104)}   & 0.403 \firstres{($\uparrow$ 0.06)}    \\
                                                                 & 720                                       & 0.407                        & 0.398                     & 0.199 \thirdres{($\downarrow$ 0.208)} & 0.281 \thirdres{($\downarrow$ 0.117)} \\
                    \midrule
                    \multirow{4}{*}{ETTh1}
                                                                 & 96                                        & 0.383                        & 0.391                     & 0.466 \firstres{($\uparrow$ 0.083)}   & 0.448 \firstres{($\uparrow$ 0.057)}   \\
                                                                 & 192                                       & 0.433                        & 0.419                     & 0.508 \firstres{($\uparrow$ 0.075)}   & 0.468 \firstres{($\uparrow$ 0.049)}   \\
                                                                 & 336                                       & 0.475                        & 0.441                     & 0.509 \firstres{($\uparrow$ 0.034)}   & 0.484 \firstres{($\uparrow$ 0.043)}   \\
                                                                 & 720                                       & 0.475                        & 0.464                     & 0.426 \thirdres{($\downarrow$ 0.049)} & 0.427 \thirdres{($\downarrow$ 0.037)} \\
                    \midrule
                    \multirow{4}{*}{ETTh2}
                                                                 & 96                                        & 0.288                        & 0.337                     & 0.386 \firstres{($\uparrow$ 0.098)}   & 0.401 \firstres{($\uparrow$ 0.064)}   \\
                                                                 & 192                                       & 0.375                        & 0.392                     & 0.432 \firstres{($\uparrow$ 0.057)}   & 0.441 \firstres{($\uparrow$ 0.049)}   \\
                                                                 & 336                                       & 0.423                        & 0.434                     & 0.466 \firstres{($\uparrow$ 0.043)}   & 0.467 \firstres{($\uparrow$ 0.033)}   \\
                                                                 & 720                                       & 0.434                        & 0.446                     & 0.304 \thirdres{($\downarrow$ 0.130)} & 0.352 \thirdres{($\downarrow$ 0.094)} \\
                    \midrule
                    \multirow{4}{*}{Electricity}
                                                                 & 96                                        & 0.197                        & 0.273                     & 0.270 \firstres{($\uparrow$ 0.073)}   & 0.363 \firstres{($\uparrow$ 0.090)}   \\
                                                                 & 192                                       & 0.197                        & 0.276                     & 0.281 \firstres{($\uparrow$ 0.084)}   & 0.371 \firstres{($\uparrow$ 0.095)}   \\
                                                                 & 336                                       & 0.212                        & 0.292                     & 0.322 \firstres{($\uparrow$ 0.110)}   & 0.399 \firstres{($\uparrow$ 0.107)}   \\
                                                                 & 720                                       & 0.253                        & 0.325                     & 0.270 \firstres{($\uparrow$ 0.017)}   & 0.361 \firstres{($\uparrow$ 0.036)}   \\
                    \midrule
                    \multirow{4}{*}{Traffic}
                                                                 & 96                                        & 0.645                        & 0.383                     & 0.727 \firstres{($\uparrow$ 0.082)}   & 0.479 \firstres{($\uparrow$ 0.096)}   \\
                                                                 & 192                                       & 0.598                        & 0.360                     & 0.734 \firstres{($\uparrow$ 0.136)}   & 0.481 \firstres{($\uparrow$ 0.121)}   \\
                                                                 & 336                                       & 0.605                        & 0.362                     & 0.768 \firstres{($\uparrow$ 0.163)}   & 0.491 \firstres{($\uparrow$ 0.129)}   \\
                                                                 & 720                                       & 0.644                        & 0.382                     & 0.774 \firstres{($\uparrow$ 0.130)}   & 0.494 \firstres{($\uparrow$ 0.112)}   \\
                    \midrule
                    \multirow{4}{*}{Weather}
                                                                 & 96                                        & 0.182                        & 0.227                     & 0.232 \firstres{($\uparrow$ 0.05)}    & 0.269 \firstres{($\uparrow$ 0.042)}   \\
                                                                 & 192                                       & 0.227                        & 0.262                     & 0.285 \firstres{($\uparrow$ 0.058)}   & 0.305 \firstres{($\uparrow$ 0.043)}   \\
                                                                 & 336                                       & 0.279                        & 0.298                     & 0.357 \firstres{($\uparrow$ 0.078)}   & 0.35 \firstres{($\uparrow$ 0.052)}    \\
                                                                 & 720                                       & 0.352                        & 0.344                     & 0.190 \thirdres{($\downarrow$ 0.162)} & 0.237 \thirdres{($\downarrow$ 0.107)} \\
                    \midrule
                    \multirow{4}{*}{Exchange}
                                                                 & 96                                        & 0.082                        & 0.198                     & 0.188 \firstres{($\uparrow$ 0.106)}   & 0.308 \firstres{($\uparrow$ 0.110)}   \\
                                                                 & 192                                       & 0.176                        & 0.297                     & 0.340 \firstres{($\uparrow$ 0.164)}   & 0.422 \firstres{($\uparrow$ 0.125)}   \\
                                                                 & 336                                       & 0.328                        & 0.412                     & 0.963 \firstres{($\uparrow$ 0.635)}   & 0.739 \firstres{($\uparrow$ 0.327)}   \\
                                                                 & 720                                       & 0.861                        & 0.700                     & 0.092 \thirdres{($\downarrow$ 0.769)} & 0.213 \thirdres{($\downarrow$ 0.487)} \\
                    \bottomrule
                \end{tabular}
            \end{small}
        \end{threeparttable}}
    \vspace{-5pt}
\end{table}

\subsection{Interpretability Analysis}
\begin{figure*}[h]
    \centering
    \subfigure[Channel BPA]{\includegraphics[width=0.9\linewidth]{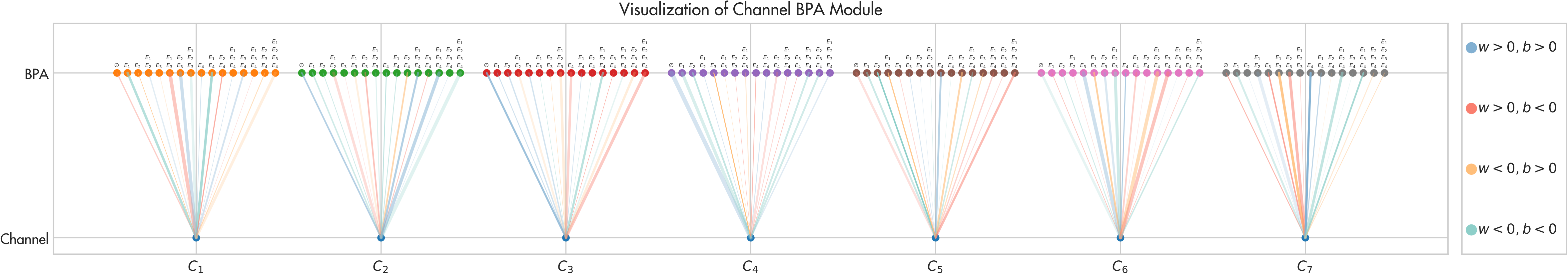}}
    \\
    \subfigure[Time BPA]{\includegraphics[width=0.9\linewidth]{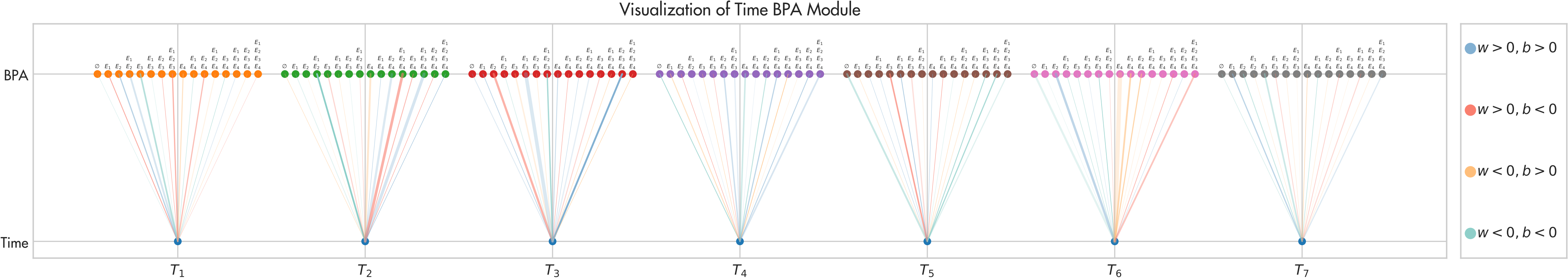}}
    \caption{Visualization of Time BPA and Channel BPA modules in target ETTh1-96: Each line represents the fuzzy set membership function corresponding to the time. The fuzzy numbers established in TEFN are all triangular fuzzy numbers, and different channels $C_i$ and time $T_i$ can correspond to fuzzy sets that conform to symbolic logic.}
    \label{fig:BPA_vis}
\end{figure*}

The Time Evidence Fusion Network (TEFN) incorporates fuzzy logic and the Dempster-Shafer theory to effectively model the inherent uncertainty and ambiguity within time series data, resulting in a model that is not only accurate but also highly interpretable. TEFN allocates mass to events within the event space $E=\{E_i|i \in [1, 2^{|S|}]\}$, where each event represents a potential outcome or scenario. The mass assigned to an event reflects the level of confidence or support for that event, providing insight into the model's belief about its likelihood of occurrence. This support allocation is particularly interpretable due to the use of fuzzy membership functions $\mu(x) = wx+b$ within the BPA module, as depicted in Figure \ref{fig:BPA_vis}. By analyzing the shape and parameters of these membership functions, users can gain a clear understanding of the model's perception of data features and their influence on predictions. For instance, it becomes clear which features are deemed more important than others, and how the model responds to variations within these features. This visualization not only makes the model's inner workings more transparent but also allows users to intuitively assess the model's confidence in its predictions and the level of uncertainty associated with each outcome.

In summary, TEFN's interpretability stems from its foundational reliance on fuzzy logic and the Dempster-Shafer theory, coupled with the transparency of its BPA and fuzzy reasoning processes. By exploring these aspects, users can develop a comprehensive understanding of the model's decision-making process and prediction outcomes, fostering trust and confidence in the model's capabilities.

\subsection{Limitation}\label{sec:lim}
TEFN is a naive and unmodified model. There is not engineering experience in the application of BPA modules, such as how to generate BPA, and fuzzy logic is not the optimal implementation method. Due to fuzzy logic, TEFN is essentially a linear model that cannot handle some nonlinear time series well. The experiment shows that adding a nonlinear activation function can effectively improve the experimental performance of TEFN. In addition, for the fusion of mass distribution in the face of the failure of the classic DSR for a large amount of data, using expectations as the fusion method may not necessarily be the optimal pattern. In terms of performance optimization, TEFN adopts a simple fully connected approach, which can lead to a rapid increase in parameter count for time series that are too long or have multiple channels. In fact, the implementation techniques of TEFN can further optimize the performance of the model through sampling and kernel operations similar to convolution. Thanks to the fact that evidence theory belongs to symbolic logic, the interpretation of neural network parameters will be more in line with human logic. Effective parameter initialization may exist, for example, initializing corresponding parameters in a non-neural network way may lead to faster convergence.

\section{Conclusion} \label{sec:con}
This paper introduces the TEFN, a novel approach for long-term time series forecasting that is particularly well-suited for handling very large datasets. TEFN treats different channels and time points as distinct information sources, using evidence theory to construct basic probability assignments and discard irrelevant data. This selective information fusion significantly reduces the computational burden, enabling TEFN to process and predict within acceptable time frames efficiently. The model's parameter efficiency, robustness, and interpretability make it a valuable tool for time series forecasting tasks involving large and complex datasets.

\bibliographystyle{IEEEtran}
\bibliography{ref}

\newpage

\begin{IEEEbiography}[{\includegraphics[width=1in,height=1.25in,clip,keepaspectratio]{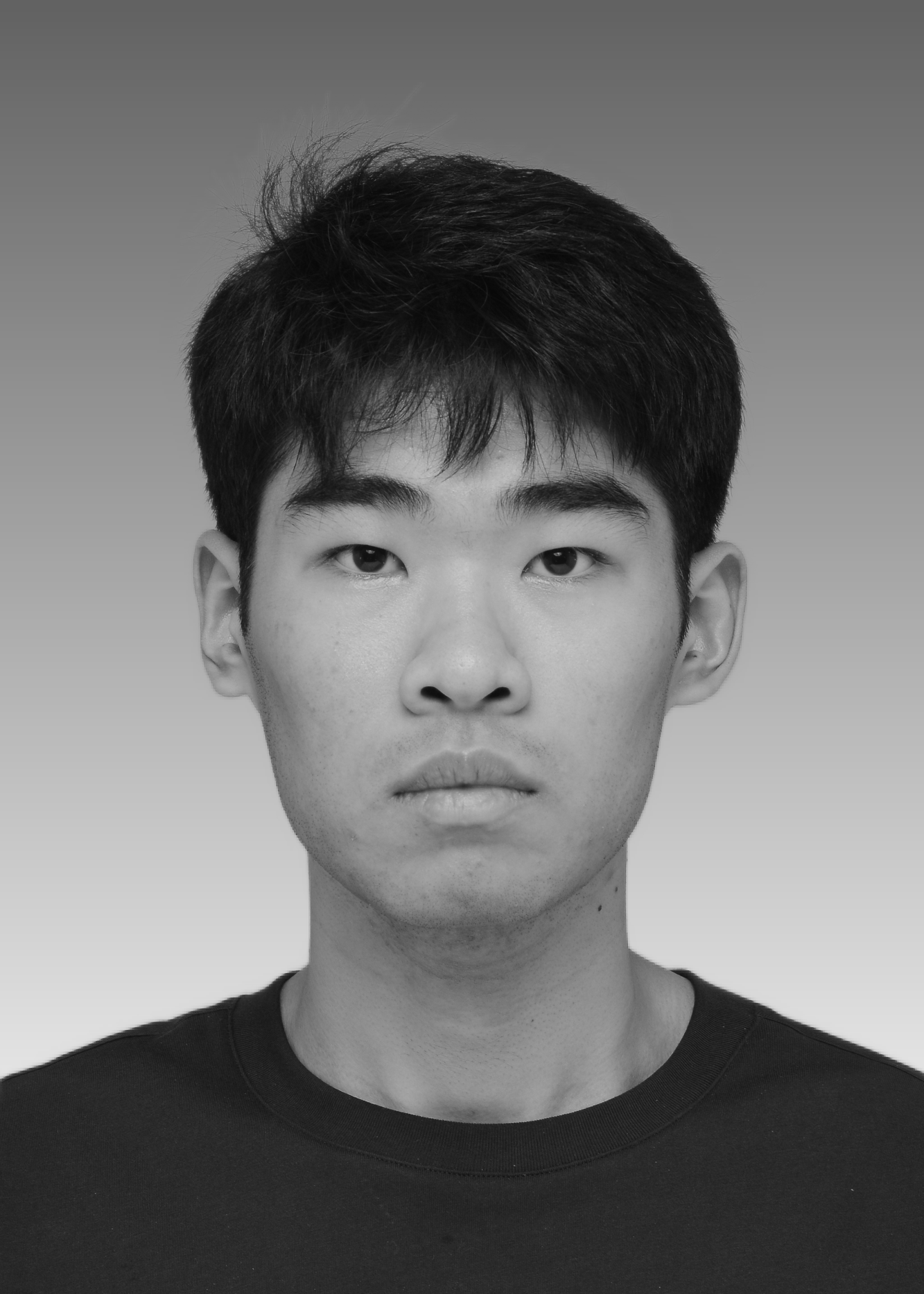}}]{Tianxiang Zhan} received the BS degree in Software Engineering from Southwest University, Chongqing, China. Currently, he is a master's student majoring in Computer Science of University of Electronic Science and Technology of China, Chengdu, China. He is also with PyPOTS Research, building AI systems for time series analysis and applications. He focuses on time series analysis, information theory, nonlinear dynamics, and complex systems.
\end{IEEEbiography}

\begin{IEEEbiography}[{\includegraphics[width=1in,height=1.25in,clip,keepaspectratio]{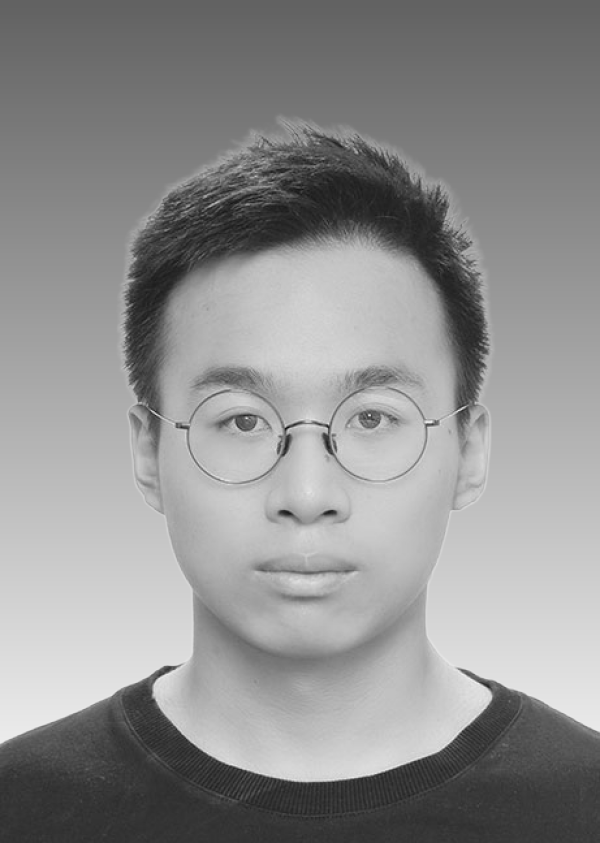}}]{Yuanpeng He} is a Ph.D candidate in Key Laboratory of High Confidence Software Technologies (Peking University), Ministry of Education, Beijing, 100871, China; School of Computer Science, Peking University, Beijing, 100871, China. His interests include computer vision and adaptive software engineering.
\end{IEEEbiography}

\begin{IEEEbiography}[{\includegraphics[width=1in,height=1.25in,clip,keepaspectratio]{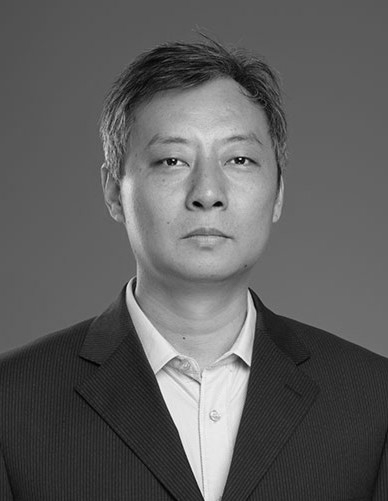}}]{Yong Deng} received the Ph.D. degree from Shanghai Jiao Tong University, Shanghai, China, in 2003. From 2005 to 2011, he was an Associate Professor with the Department of Instrument Science and Technology, Shanghai Jiao Tong University. In 2010, he was a Professor with the School of Computer and Information Science, Southwest University, Chongqing, China. In 2012, he was a Visiting Professor with Vanderbilt University, Nashville, TN, USA. In 2016, he was also a Professor with the School of Electronic and Information Engineering, Xi'an Jiaotong University, Xi'an, China. Since 2017, he has been a Full Professor with the Institute of Fundamental and Frontier Science, University of Electronic Science and Technology of China, Chengdu, China. He is a panel member of the information division of NSFC. He is a JSPS Invitational Fellow at the Japan Advanced Institute of Science and Technology, Ishikawa, Japan. He has authored or coauthored more than 100 papers in refereed journals. His research interests include uncertainty, quantum computation, and complex system modeling. He presents generalized evidence theory, D numbers, Deng entropy, information volume of mass function, and Random Permutation Set.
\end{IEEEbiography}

\begin{IEEEbiography}[{\includegraphics[width=1in,height=1.25in,clip,keepaspectratio]{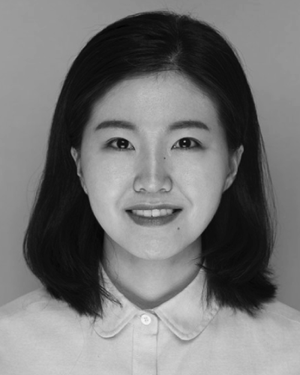}}]{Zhen Li} received the Ph.D. degree in industrial engineering and management from Peking University, Beijing, China, in 2022. She is currently an Engineer with the China Mobile Information Technology Center, Beijing. Her research interests are focused on data mining, prognosis, and health management through advanced data analytics, quality, and reliability engineering.
\end{IEEEbiography}

\begin{IEEEbiography}[{\includegraphics[width=1in,height=1.25in,clip,keepaspectratio]{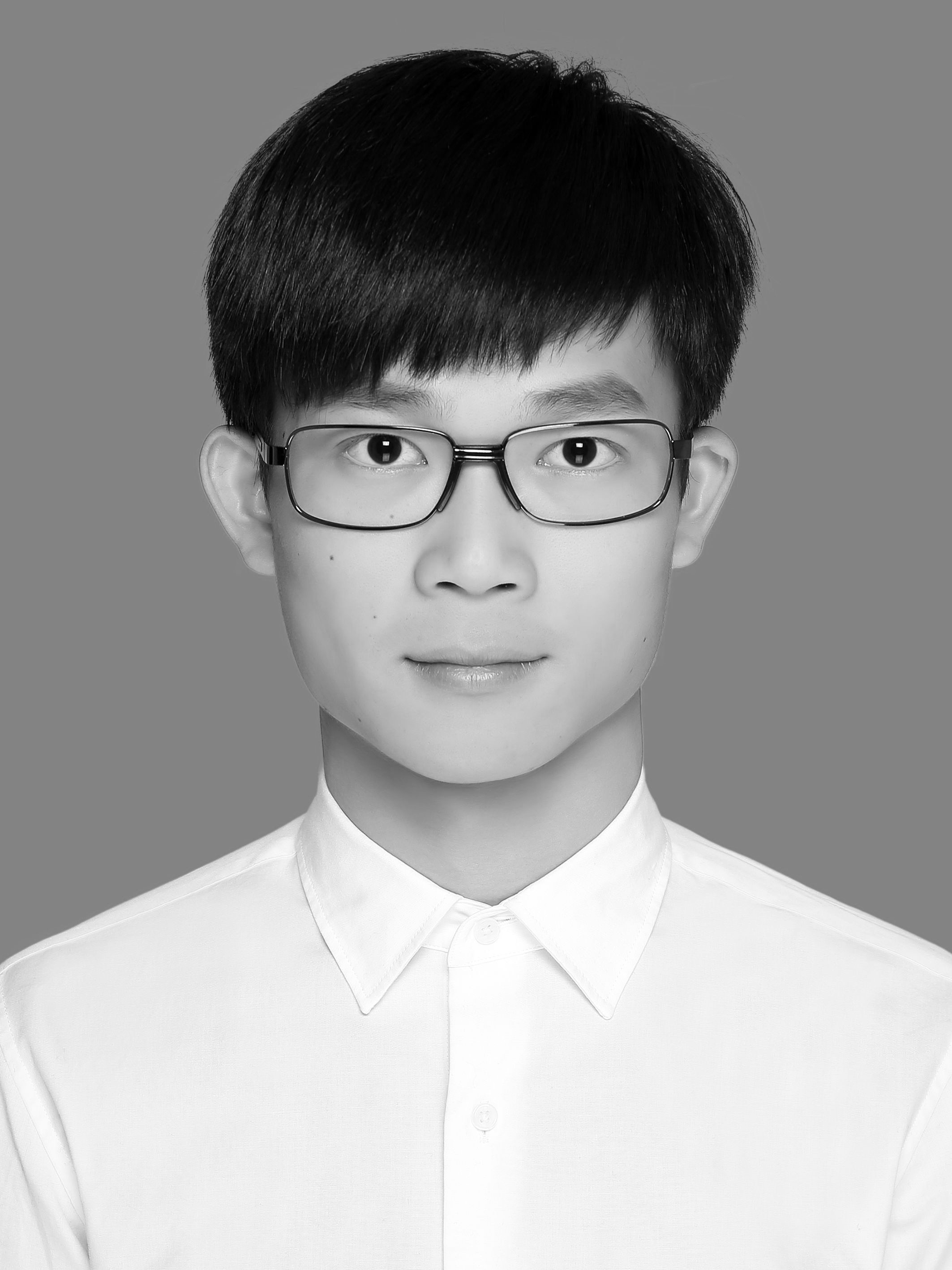}}]{Wenjie Du} received the Master of Applied Science in Electrical and Computer Engineering from Concordia University, Montreal, Canada. His research majors in modeling time series with machine learning, especially partially-observed time series (POTS). He is currently with PyPOTS Research, building AI systems for time series analysis and applications.
\end{IEEEbiography}

\newpage

\begin{IEEEbiography}[{\includegraphics[width=1in,height=1.25in,clip,keepaspectratio]{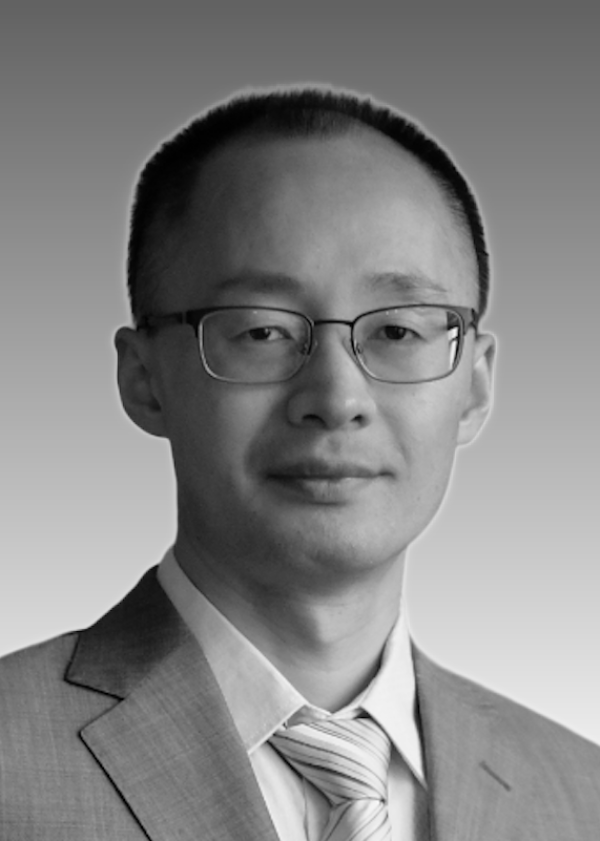}}]{Qingsong Wen} is currently the Head of AI \& Chief Scientist at Squirrel Ai Learning. Before that, he worked at Alibaba, Qualcomm, Marvell, etc., and received his M.S. and Ph.D. degrees in Electrical and Computer Engineering from Georgia Institute of Technology, USA. His research interests include machine learning, data mining, and signal processing, especially AI for Time Series, AI for Education, LLM \& AI Agent. He has published around 150 top-ranked AI conference and journal papers, had multiple Oral/Spotlight Papers at NeurIPS, ICML, ICLR, ACL, AAAI, had multiple Most Influential Papers at IJCAI, received multiple IAAI Innovative Application Awards at AAAI, and won First Place of SP Grand Challenge at ICASSP. Currently, he serves as Chair of IEEE CIS Task Force on AI for Time Series and Spatio-Temporal Data, and Vice Chair of INNS AI for Education Section. He also regularly serves as Area Chair of the top conferences including NeurIPS, ICML, KDD, IJCAI, ICASSP, etc. In addition, he serves as Associate Editor for IEEE Transactions on Pattern Analysis and Machine Intelligence, IEEE Signal Processing Letters, and Neurocomputing.
\end{IEEEbiography}

\vfill

\end{document}